\pdfoutput=1

\documentclass[11pt]{article}

\usepackage[]{latex/acl}

\usepackage{times}
\usepackage{latexsym}

\usepackage[T1]{fontenc}

\usepackage[utf8]{inputenc}

\usepackage{microtype}

\usepackage{inconsolata}

\usepackage{amssymb}%
\usepackage{pifont}%
\newcommand{\cmark}{\ding{51}}%
\newcommand{\xmark}{\ding{55}}%
\usepackage[inline,shortlabels]{enumitem}
\usepackage{amsmath}
\usepackage{amsfonts}
\usepackage{comment}
\usepackage{multirow}

\usepackage{hyperref}
\usepackage{todonotes}
\usepackage{microtype}
\usepackage{xcolor}
\usepackage{tikz,lipsum}
\usepackage{kotex}
\usepackage{soul}
\usepackage{array, booktabs, multirow}
\usepackage{color, colortbl} 
\usepackage{arydshln} 
\usepackage{enumitem} %
\usepackage{subcaption} %
\usepackage{graphicx} %
\usepackage{tcolorbox,multicol} %
\usepackage{cuted} %
\usepackage{xspace}
\usepackage[export]{adjustbox} %
\usepackage{bm} %

\definecolor{LightGray}{gray}{0.9}

\NewTColorBox{NewBox}{ s O{!htbp} }{%
  floatplacement={#2},
  IfBooleanTF={#1}{float*,width=\textwidth}{float},
  colframe=blue!50!black,colback=blue!10!white,%
  }
\newcommand{\lexcgen}{\xspace{\fontfamily{qcs}\selectfont LexC-Gen}\xspace}

\usepackage{cleveref}
\crefname{algorithm}{Alg.}{Algs.}
\Crefname{algorithm}{Algorithm}{Algorithms}
\crefname{appendix}{App.}{App.}
\Crefname{appendix}{Appendix}{Appendices}
\crefname{figure}{Fig.}{Figs.}
\Crefname{figure}{Figure}{Figures}
\Crefname{section}{Section}{Sections}
\crefname{section}{Sect.}{Sect.}
\crefname{subsection}{Sect.}{Sect.}
\Crefname{subsection}{Section}{Sections}
\crefname{subsubsection}{Sect.}{Sect.}
\Crefname{subsubsection}{Section}{Sections}
\crefname{table}{Table}{Tables}
\Crefname{table}{Table}{Tables}

\title{\lexcgen: Generating Data for Extremely Low-Resource Languages \\with Large Language Models and Bilingual Lexicons}

\author{
\textbf{Zheng-Xin Yong}$^{1}$ \quad
\textbf{Cristina Menghini}$^{2}$ \quad
\textbf{Stephen H. Bach}$^{1}$
\\
$^{1}$Department of Computer Science, Brown University \\
$^{2}$Data Science Institute, Brown University \\
\texttt{\{contact.yong,cristina\_menghini,stephen\_bach\}@brown.edu}
}

\begin{document}
\maketitle
\begin{abstract}

Data scarcity in low-resource languages can be addressed with word-to-word translations from labeled task data in high-resource languages using bilingual lexicons. 
However, bilingual lexicons often have limited lexical overlap with task data, which results in poor translation coverage and lexicon utilization.
We propose lexicon-conditioned data generation (\lexcgen), a method that generates low-resource-language classification task data at scale. Specifically, \lexcgen first uses high-resource-language words from bilingual lexicons to generate lexicon-compatible task data, and then it translates them into low-resource languages with bilingual lexicons via word translation.
Across 17 extremely low-resource languages, \lexcgen generated data is competitive with expert-translated gold data, and yields on average 5.6 and 8.9 points improvement over existing lexicon-based word translation methods on sentiment analysis and topic classification tasks respectively.
Through ablation study, we show that conditioning on bilingual lexicons is the key component of \lexcgen. \lexcgen serves as a potential solution to close the performance gap between open-source multilingual models such as BLOOMZ and state-of-the-art commercial models like GPT-4o on low-resource-language tasks.

\end{abstract}

\section{Introduction}
\textit{Extremely low-resource languages} do not have any labeled data and are thereby considered the ``Left-Behinds'' in NLP language technology development \citep{joshi-etal-2020-state,mabokela2022multilingual,robinson-etal-2023-chatgpt}. Nonetheless, many of them have \textit{bilingual lexicons} resources, which are usually the first product of language documentation \citep{meara1993bilingual,schreuder1993bilingual,kroll2017bilingual}. Bilingual lexicons are dictionaries that map words from one language to their translations in another languages, and they cover more than 5000 languages around the world \cite{wang-etal-2022-expanding,koto2024lexicon}.

\begin{figure}[!t]
    \centering
    \begin{subfigure}{0.22\textwidth}
        \includegraphics[width=\textwidth]{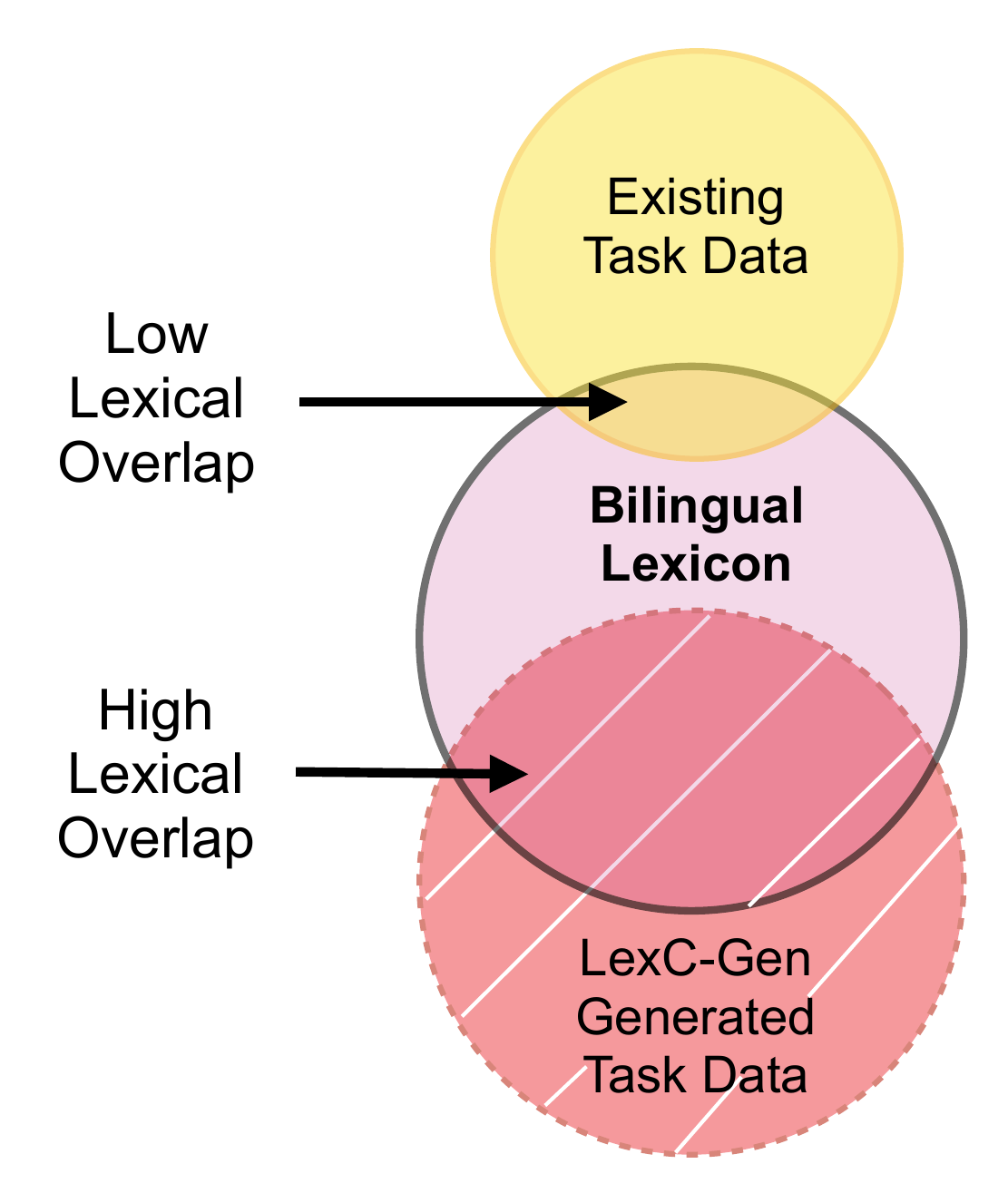}
        \caption{Intuition of \lexcgen.}
        \label{fig:lexcgen-intuition}
    \end{subfigure}
    \hfill
    \begin{subfigure}{0.22\textwidth}
        \includegraphics[width=\textwidth]{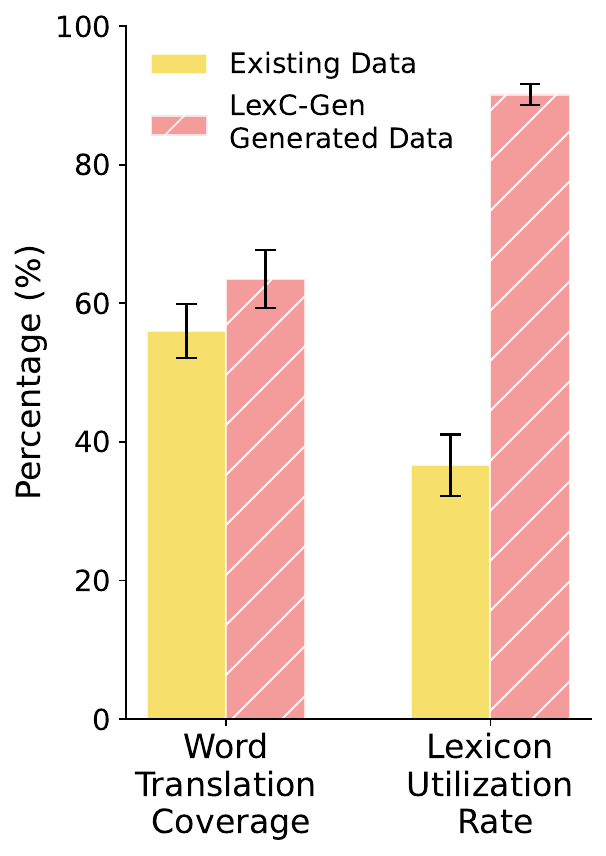}
        \caption{Lexical overlap between lexicons and data.}
        \label{fig:lexcgen-coverage-utilrate}
    \end{subfigure}
    \caption{
    We observe data-lexicon mismatch (i.e., low lexical overlap) between existing task data and bilingual lexicons (\Cref{fig:lexcgen-intuition}).
    \lexcgen addresses the issue by generating data using words from lexicons so the data will have more words translated (i.e., higher word translation coverage) and higher lexicon utilization rate (\Cref{fig:lexcgen-coverage-utilrate}).} 
    \label{fig:lexcgen-intro}
    \vspace{-5mm}
\end{figure}

\begin{figure*}[!t]
    \centering\includegraphics[width=\textwidth]{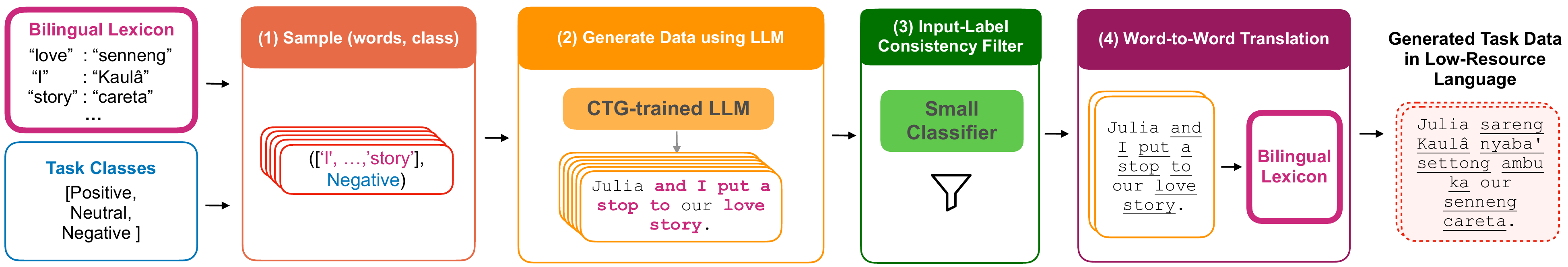}
    \caption{\textbf{\lexcgen} Given a bilingual lexicon and the set of classes for a classification task, (1) we randomly sample the class label and a set of words from bilingual lexicon, for as many instances we desire to generate. (2) We use these pairs to build the prompts to query CTG-trained LLM (\Cref{fig:ctg}) and generate the task data in high-resource language. (3) Then, we train a task classifier on existing task data to filter generated data and ensure input-label consistency. (4) After filtering, we apply word-to-word translation with the bilingual lexicon following prior work~\citep{wang-etal-2022-expanding}. Finally we get the synthetic task data for the target low-resource language, which is used to finetune task classifier.}
    \label{fig:lexcgen-method}
    \vspace{-3mm}
\end{figure*}

Previous work uses bilingual lexicons to directly translate labeled data from high-resource languages to low-resource languages through word-for-word substitution \citep[inter alia]{wang-etal-2022-expanding, jones-etal-2023-gatitos}. However, we argue that it is ineffective because of \textit{data-lexicon mismatch}. Often, the words in the \textit{existing task data}---readily available labeled data in high-resource languages for a target task, e.g., sentiment analysis or topic classification---have low lexical overlap with the words in the task-agnostic bilingual lexicons, as shown in~\Cref{fig:lexcgen-intro}. This mismatch not only results in many words remain untranslated, but also causes entries in the bilingual lexicon, which possibly contain useful semantic information for downstream tasks, missing from the translated dataset.

In this work, we introduce \textbf{\lexcgen},\footnote{pronounced as lek-see-jen} which is a \textbf{lex}icon-\textbf{c}onditioned data \textbf{gen}eration method, to mitigate data-lexicon mismatch through synthetic data generation.  
Specifically, we train LLMs to generate task data using words from bilingual lexicons, so the data have a higher lexical overlap with the lexicons.
This results in better word translation coverage and lexicon utilization rate (\Cref{fig:lexcgen-intro}). 
We also propose a quality-control method that checks for input-label consistency to filter out poor-quality generated data.

We evaluated \lexcgen across 17 extremely low-resource languages on sentiment analysis and topic classification tasks. We found that finetuning classifiers on \lexcgen generated data improves on average 5.6 and 8.9 points in accuracy respectively over word-translated existing training data \citep{wang-etal-2022-expanding}. 
Surprisingly, finetuning on \lexcgen word-translated data even matches the performance of finetuning on \textit{gold data} in the target language curated by native speakers or professional translators. We show that lexicon-conditioning is the critical success factor of \lexcgen.

Finally, we discuss how \lexcgen helps close the performance gap of open-source LLMs in low-resource-language tasks. We show that instead of zero-shot or few-shot prompting, it is better to use them to generate training data with \lexcgen. The data generation process is cost-effective, and the permissive nature of the models allows generated data to be made open access for further research and building systems for extremely low-resource languages, which benefits multilingual NLP progress for these data-scarce languages.

Our contributions can be summarized as follows:

\setlist{nolistsep}
    \begin{enumerate}[noitemsep]
        \item We present \lexcgen, a method that conditions LLMs on bilingual lexicons to generate low-resource-language task data to address \textit{data-lexicon mismatch} problem.
        \item We demonstrate that training on word-translated task data can match training on \textit{gold data} for extremely low-resource-languages.
        \item Our extensive ablation study on \lexcgen shows that simply scaling up generated task data is \textit{insufficient}. Lexicon-conditioning is necessary to maximize lexical overlap between task data and bilingual lexicons.
    \end{enumerate}

\section{Related Work}

\paragraph{Generating task data with LLMs}
LLM-powered data generation is a recent promising area of research that enables cost-effective collection of diverse task data with minimal human labor \citep{honovich-etal-2023-unnatural,radharapu2023aart,wang-etal-2023-self-instruct,nayak2023bonito,yehudai2024genie}. Nonetheless, this line of work has been underexplored in a multilingual setting. \citet{whitehouse-etal-2023-llm} demonstrated that GPT-4's generated multilingual training data for commonsense reasoning task in mid-/high-resource languages can improve cross-lingual performance. However, language coverage of LLMs and translation models are significantly smaller than lexicons \citep{wang-etal-2022-expanding,bapna2022building, koto2024lexicon}. Instead, we use LLMs to generate task data that maximize lexical overlap with bilingual lexicons for translations, and we show that our synthetic data can improve NLU semantic task performance in extremely low-resource languages. 

\paragraph{Lexicon-based cross-lingual data augmentation}
Lexicon-based augmentation creates data for low-resource languages by swapping words in high-resource-language data with their dictionary word translations in bilingual lexicons. This is useful for low-resource languages that cannot be readily translated by translation models/APIs with limited language coverage. Prior work has demonstrated their effectiveness across a wide range of NLP tasks, such as machine translation \citep{streiter2000learning,ramesh-sankaranarayanan-2018-neural,thompson-etal-2019-hablex,kumar2022dict,jones-etal-2023-gatitos}, sequence labeling \citep{scherrer2013lexicon,mayhew-etal-2017-cheap,wang-etal-2022-expanding}, sentiment classification \citep{rasooli2018cross,ali-etal-2021-creating,mohammed2023building}, and topic classification \cite{song2019toward}. However, many lexicon-based data augmentation strategies for semantic tasks in low-resource languages rely on domain-specific lexicons \cite{das2010sentiwordnet,buechel-etal-2016-feelings,mohammed2023building,koto2024lexicon}, and performance-wise they still fall short of gold training data collected in the target low-resource language \cite{rasooli2018cross,koto2024lexicon}. Our method \lexcgen not only works with domain-agnostic bilingual lexicons, but also demonstrates competitive performance with gold training data on sentiment analysis and topic classification tasks across many low-resource languages.

\setul{1.5pt}{.4pt}
\begin{table*}[ht]
    \small
   \centering
   \begin{tabular}{lccccccccc}
   \toprule
   \textbf{Methods} & \textbf{\#data} & \textbf{ace} & \textbf{ban} & \textbf{bbc} & \textbf{bjn} & \textbf{bug} & \textbf{mad} & \textbf{min} & \textbf{Avg}  \\ 
   \midrule

   \emph{Zero-shot/Few-shot prompting}\\
   \midrule
   BLOOMZ-7.1.B & 0 & 47.0 & 50.5 & 43.0 & 49.5 & 38.5 & 48.0 & 52.5 & 47.0 \\
       
    Aya-101-13B & 0 & 58.8 & 59.2 & 48.1 & \ul{82.8} & 35.9 & 48.4 & \ul{77.9} & 58.7\\
    Aya-101-13B (few-shot) & 5 & 60.8 & \textbf{62.6} & 53.0 & \textbf{83.9} & 45.7 & 53.9 & \textbf{79.9} & 62.8 \\
    \rowcolor{LightGray} GPT-4o & 0 & 75.3 & 81.3 & 65.8 & 83.8 & 51.5 & 74.0 & 85.3 & 73.8
    \\
     \midrule
     \emph{Cross-lingual zero-shot} \\
     \midrule
   Existing Task Data (\texttt{en}) & 500 & 56.8 & 60.2 & 51.1 & 63.3 & 45.8 & 56.0 & 57.7 & 55.8 \\
    \midrule
   \emph{Word translation} \\
   \midrule
   Existing Task Data (\texttt{T}) & 500 & 63.6 & 58.3 & 55.8 & 66.4 & 57.7 & 59.3 & 71.6 & 61.8 \\

   \quad + Existing Task Data (\texttt{en}) & 1000 & 67.8 & 62.4 & 60.4 & 66.3 & 56.7 & {62.4} & 75.1 & 64.4 \\
   
   \quad + Label Distillation & \multirow{2}{*}{1000} &
   \multirow{2}{*}{58.8} & \multirow{2}{*}{52.9} & \multirow{2}{*}{45.7} & \multirow{2}{*}{58.8} & \multirow{2}{*}{43.9} & \multirow{2}{*}{56.8} & \multirow{2}{*}{68.7} & \multirow{2}{*}{55.1} \\
   \quad \ \ \citep{wang-etal-2022-expanding} \\

   \hdashline\noalign{\vskip 0.5ex}
   \lexcgen-1K (\texttt{T}) & $\sim370$ & 42.4 & 47.1 & 49.6 & 53.9 & 43.5 & 42.3 & 44.3 & 46.2 \\
   \quad + Existing Task Data (\texttt{en}) & $\sim870$ & 67.8 & 62.4 & 60.4 & 66.3 & 56.7 & {62.4} & 75.1 & 64.4 \\

   \lexcgen-10K (\texttt{T}) & $\sim3.7$K & 66.6 & 67.1 & 61.0 & 72.3 & 57.3 & 61.2 & 70.7 & 65.2 \\
   \quad + Existing Task Data (\texttt{en}) & $\sim4.2$K & 68.2 & 67.0 & 62.8 & 71.4 & 58.5 & \ul{57.9} & 70.3 & 65.2 \\
   
   \textbf{\lexcgen-100K} (\texttt{T}) & $\sim37$K & \ul{70.0} & {71.5} & \ul{65.1} & 73.4 & \ul{63.7} & \textbf{69.9} & 76.5 & \ul{70.0} \\
   \quad \textbf{+ Existing Task Data (\texttt{en})} & $\sim38$K & \textbf{70.7} & \ul{71.4} & \textbf{67.8} & {74.6} & \textbf{65.8} & \textbf{69.9} & {76.9} & \textbf{71.0} \\
   \midrule
    \rowcolor{LightGray}\emph{Gold Translations} & 500 & 72.1 & 71.6 & 68.6 & 72.8 & 68.1 & 66.7 & 77.3 & 71.0 \\
   \bottomrule
   \end{tabular}
    \caption{Sentiment analysis accuracy on 7 Indonesian extremely low-resource local languages in the NusaX dataset \cite{winata-etal-2023-nusax}. (\texttt{T}) indicates word-translated data, and (\texttt{en}) refers to the existing task data in English. The terms -1K, -10K and -100K refer to the size of training data generated by \lexcgen before filtering. 
    We \textbf{bold} the best result and \ul{underline} the second-best. We report results averaged over 5 seeds of classifier finetuning.}
   \label{tab:sa-result}
   \vspace{-2.5mm}
\end{table*}

\vspace*{-.2mm} %
\section{\lexcgen}
We aim to generate data for classification tasks in a low-resource language $L$, given access to 
(1) labeled task data $\mathcal{T}_H$ with $C$ classes in a high-resource language $H$, 
(2) a bilingual lexicon $D^L_H$ that maps words from $H$ to $L$, and 
(3) an LLM supporting $H$. 

\lexcgen uses these inputs to generate
labeled task data $\widetilde{\mathcal{T}}_L$ in low-resource language. 
Our key idea is to prompt the LLM to generate task data using high-resource-language words from bilingual lexicons in order to create task data that have a higher lexical overlap with those bilingual lexicons (\Cref{fig:lexcgen-intuition}), and thus can be more effectively translated into $L$. In the following, we describe the steps to obtain $\widetilde{\mathcal{T}}_L$. 
For readability, we refer to $D^L_H$ as $D$.

\subsection{Sample Lexicon Words and Class Label}\label{sec:method-sample}
First, we randomly sample a set $W_H$ of high-resource-language words $w_H$ from $D$ and a class label $c$. This corresponds to step (1) in \Cref{fig:lexcgen-method}. The goal is to prompt our LLM to generate task inputs of class $c$ using as many words from $W_H$ as possible.

\subsection{Generate Data with LLM Trained with Controlled-Text Generation (CTG)}\label{sec:method-generate-data}
Next, we prompt an LLM to generate high-resource-language task data $\widetilde{\mathcal{T}}_{H|D}$ conditioned on the bilingual lexicon. This is step (2) in \Cref{fig:lexcgen-method}. 
However, because open-access instruction-tuned LLMs such as BLOOMZ \citep{muennighoff-etal-2023-crosslingual} are not finetuned for this purpose, we carry out controlled text generation (CTG) training of LLMs \citep{zhang2023survey, zhou23ctg} to create CTG-trained LLM.

\begin{figure}[t]
  \centering
  \includegraphics[width=0.45\textwidth]{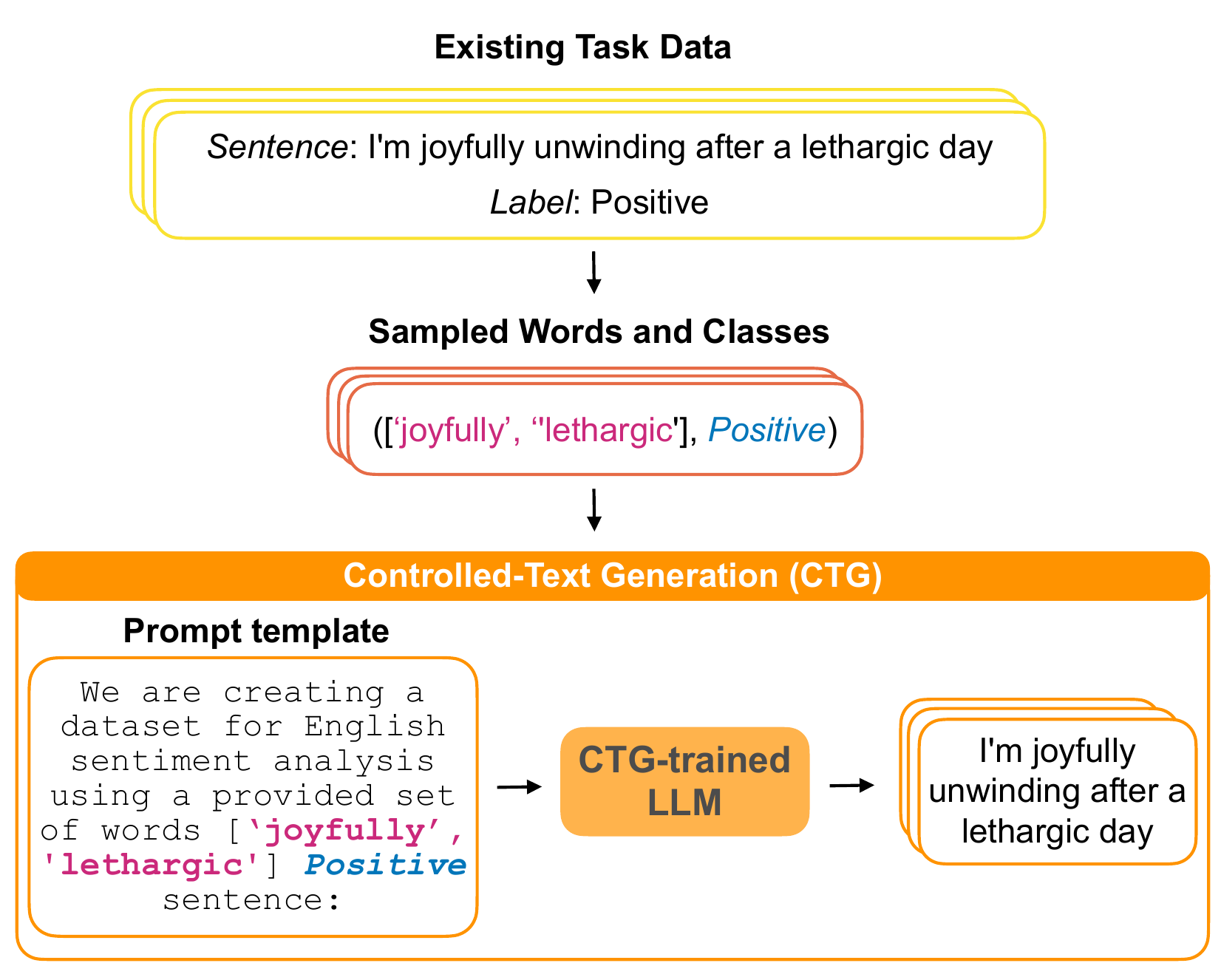}
  \caption{\textbf{Controlled-Text Generation (CTG) training.} This figure shows the pipeline for the LLM finetuning for CTG. We construct the training data starting from the existing labeled task data $\mathcal{T}_H$. From each instance $t_H$, we sample without replacement a set of words $W_H$ and associate it to class $c$. This information is plugged into the prompt template, and it is used to finetune an LLM that generates sentences conditioned on $c$ and $W_H$.}
  \label{fig:ctg}
\end{figure}

\paragraph{CTG Training} We construct CTG training data from existing task data $\mathcal{T}_H$. 
Each instance $t_H \in \mathcal{T}_H$ consists of a pair of text $x_H$ and task label $c$. 
We randomly sample a variable number of word tokens $w_H$ uniformly at random without repetition from $x_H$ to create $W_H$. 
Then, we format the CTG training data using the prompt template in~\Cref{fig:ctg}, so that the LLM learns to generate task input $\tilde{x}_{H|c, W_H}$ conditioned on $c$ and $W_H$. 

CTG training is data-efficient. We found that generating only a single CTG training example per each $t_H \in \mathcal{T}_H$ is already sufficient to instruction-tune the model. Specifically, our CTG training data consists of 500 and 701 instances for our sentiment analysis and topic classification tasks respectively.

\paragraph{Task Data Generation} After CTG training, we prompt the LLM reusing the template in~\Cref{fig:ctg}, but now we use lexicon words with random task class labels from \Cref{sec:method-sample}. We can now generate synthetic high-resource-language task data  $\widetilde{\mathcal{T}}_{H|D}$ at scale conditioned on bilingual lexicons.

\subsection{Input-Label Consistency Filter}\label{sec:method-filter}
To ensure high-quality data, we apply an input-label consistency filter after data generation to reduce training noise from labeling errors.
For instance, CTG-trained LLM may generate a sentence with negative sentiment even though the specified task label $c$ is positive sentiment in the input prompt (\Cref{fig:ctg}).
Therefore, we finetune a small classifier mBERT on the same existing task data $\mathcal{T}_H$ and use it to relabel $\widetilde{\mathcal{T}}_{H|L}$. 
Then, we filter out all data instances where the classifier's prediction does not match the generated input-label pairs. 

At this point (step (3) in \Cref{fig:lexcgen-method}), we have high-quality lexicon-compatible task data in language $H$ that allows for better word-to-word translation into language $L$ by using $D$.

\subsection{Word-to-Word Translation into Low-Resource Languages} \label{sec:method-translate}
Finally, we carry out word-to-word translation following the procedures in prior work \citep{wang-etal-2022-expanding,jones-etal-2023-gatitos}. 
We use $D$ to substitute the high-resource-language words $w_H \in \widetilde{\mathcal{T}}_{H|D}$ with their low-resource-language word translation $w_L$, thus creating $\widetilde{\mathcal{T}}_L$. 
We randomly sample $w_L$ if $w_H$ has multiple possible translations and keep $w_H$ as is in $\widetilde{\mathcal{T}}_{H|D}$ if there is no translation for it in $D$. 
After we obtain the synthetic cross-lingual task data $\widetilde{\mathcal{T}}_L$, we use it as training data to finetune a classifier for the target task in the low-resource-language.

\setul{1.5pt}{.4pt}
\begin{table*}[ht]
    \small
   \centering
   \begin{tabular}{lcccccccccccc}
   \toprule
   \textbf{Methods} & \textbf{\#data} & \textbf{bam} & \textbf{ewe} & \textbf{fij} & \textbf{grn} & \textbf{lin} & \textbf{lus} & \textbf{sag} & \textbf{tso} & \textbf{tum} & \textbf{twi} & \textbf{Avg }\\ 
   \midrule

   \emph{Zero-/Few-shot prompting}\\
   \midrule
   BLOOMZ-7.1.B & 0 & 41.7 & 34.3 & 35.3 & 41.7 & 42.2 & 38.7 & 36.8 & 41.7 & 40.2 & 41.7 & 39.4\\
   
    Aya-101-13B & 0 & 36.8 & 39.1 & 50.9 & 48.8 & 52.4 & 43.7 & 40.2 & \ul{54.1} & 50.0 & 37.7 & 45.4 \\
    Aya-101-13B (few-shot) & 5 & 42.2 & 46.1 & 60.4 & 55.1 & 59.7 & 48.2 & 49.4 & \textbf{56.2} & \textbf{57.5} & 43.8 & 51.9 \\
    \rowcolor{LightGray} GPT-4o & 0 & 58.1 & 56.2 & 63.9 & 75.8 & 69.4 & 65.3 & 57.8 & 57.2 & 59.8 & 64.8 & 67.7 \\
    
     \midrule
     \emph{Cross-lingual zero-shot} \\
     \midrule
   Existing Task Data (\texttt{en}) & 701 & 29.6 & 32.5 & 42.5 & 57.7 & 42.0 & 49.9 & 37.6 & 39.6 & 40.3 & 40.7 & 41.2 \\
    \midrule
   \emph{Word translation} \\
   \midrule
   Existing Task Data (\texttt{T}) & 701 & 40.2 & 41.4 & 49.1 & 63.9 & 52.3 & 61.8 & 46.7 & 39.1 & 42.5 & 54.9 & 49.2 \\

   \quad + Existing Task Data (\texttt{en}) & 1402 & 42.5 & 41.4 & 47.8 & 67.2 & 55.9 & 63.4 & 47.9 & 40.0 & 43.4 & 56.4 & 50.6
   \\
   
   \quad + Label Distillation & \multirow{2}{*}{1402} & \multirow{2}{*}{37.5} & \multirow{2}{*}{33.1} & \multirow{2}{*}{41.9} & \multirow{2}{*}{59.0} & \multirow{2}{*}{37.8} & \multirow{2}{*}{56.5} &  \multirow{2}{*}{38.5} & \multirow{2}{*}{42.1} & \multirow{2}{*}{41.2} & \multirow{2}{*}{35.0} & \multirow{2}{*}{42.3} \\
   \quad \ \ \citep{wang-etal-2022-expanding} \\
   
    \hdashline\noalign{\vskip 0.5ex}
   \lexcgen-1K (\texttt{T}) & $\sim220$ & 22.9 & 37.8 & 40.2 & 50.1 & 45.0 & 52.5 & 40.9 & 29.2 & 37.6 & 42.1 & 39.8 \\
    \quad + Existing Task Data (\texttt{en}) & $\sim920$ & 36.5 & 41.2 & 45.3 & 68.3 & 53.0 & 61.9 & 49.1 & 37.1 & 39.0 & 53.7 & 48.5
   \\
   
   \lexcgen-10K (\texttt{T}) & $\sim2.2$K & 38.5 & 40.5 & 51.4 & 67.1 & 57.6 & 64.1 & 55.3 & 41.1 & 42.6 & 55.1 & 51.3 \\
   \quad + Existing Task Data (\texttt{en}) & $\sim2.9$K & 33.8 & 42.6 & 51.3 & 67.1 & 59.3 & 64.8 & 53.8 & 43.8 & 43.2 & 54.3 & 51.4
   \\
   
   \textbf{\lexcgen-100K} (\texttt{T}) & $\sim22$K & \ul{44.0} & \textbf{51.1} & \textbf{70.2} & \textbf{74.3} & \textbf{67.4} & \textbf{69.3} & \textbf{61.0} & 42.2 & 50.9 & \ul{64.9} & \textbf{59.5} \\
   \quad + Existing Task Data (\texttt{en}) & $\sim23$K & \textbf{46.2} & \ul{47.6} & \ul{68.0} & \ul{73.0} & \ul{67.2} & \ul{68.9} & \ul{57.0} & {42.6} & \ul{53.0} & \textbf{65.8} & \ul{58.9}
   \\

   \midrule
    \rowcolor{LightGray}\emph{Gold Translations} & 701 & 54.9 & 53.0 & 61.7 & 71.2 & 64.6 & 68.4 & 60.7 & 55.9 & 63.4 & 62.2 & 61.6\\
   \bottomrule
   \end{tabular}
    \caption{Topic classification accuracy for 10 worst-performing languages in the SIB-200 dataset \cite{adelani2023sib200}. We follow the schema defined in Table~\ref{tab:sa-result}. 
    }
   \label{tab:tc-result}
   \vspace{-2.5mm}
\end{table*}

\section{Experimental Setup}

We compare \lexcgen against baselines and gold translations on sentiment analysis and topic classification tasks. We describe the task datasets in \Cref{sec:setup-task-dataset}, how we instantiate \lexcgen in \Cref{sec:setup-lexcgen}, and our baselines as well as gold translations in \Cref{sec:setup-baseline}.

\subsection{Tasks and Datasets}\label{sec:setup-task-dataset}
We evaluate \lexcgen on sentiment analysis and topic classification tasks across 17 low-resource languages. The task datasets contain \textit{gold training data} that are curated with translations by native speakers or professional translators. Detailed information for the tasks and languages can be found in \Cref{app:task-languages}. 

\paragraph{Sentiment analysis}
We use the NusaX sentiment analysis dataset \citep{winata-etal-2023-nusax} developed for Indonesian low-resource languages. The dataset has 3 sentiment labels: positive, neutral, and negative.
In our setup, we evaluate~\lexcgen on 7 languages that also exist in the Gatitos lexicon. 

\paragraph{Topic classification} 
SIB-200~\citep{adelani2023sib200} is a topic classification benchmark that covers 200 languages and 7 topic categories.
We evaluate~\lexcgen on the \textit{10 worst-performing languages} that we found to have the largest performance gap between gold translations and the word translation baseline \cite{wang-etal-2022-expanding}.

\subsection{\lexcgen Instantiation}\label{sec:setup-lexcgen}

\paragraph{LLM}
We use the BLOOMZ model \citep{muennighoff-etal-2023-crosslingual} with 7.1 billion parameters (BLOOMZ-7.1B) as our initial instruction-tuned LLM. This allows us to compare performance between its zero-shot prompting and its usage with \lexcgen. 

\paragraph{Bilingual lexicons} 
We choose Gatitos bilingual lexicons \citep{jones-etal-2023-gatitos} to translate the generated English data into low-resource languages. Gatitos includes English entries such as frequent English words, numbers, and time, and they are translated into 170 extremely low-resource languages. Gatitos have been manually reviewed, so its entries have higher quality than other bilingual lexicons such as Panlex \citep{kamholz2014panlex}.

\paragraph{Generated task data}
We first use \lexcgen to generate English datasets with 1K, 10K, and 100K instances, to which we refer as \lexcgen-1K, \mbox{-10K}, and \mbox{-100K}, before filtering out mismatched input-label pairs. The effective data size after filtering with input-label consistency checking is between 20\% and 40\% of the generated task data. Then, we use Gatitos lexicons \citep{jones-etal-2023-gatitos} to translate them into low-resource languages.

\paragraph{Training and data generation with LLM}
We provide further training and inference details of \lexcgen in \Cref{app:lexcgen-details}. We also showcase examples of the generated data for sentiment analysis in \Cref{tab:sa-text-samples} and for topic classification in \Cref{tab:tc-text-samples}.

\begin{table*}[h]
    \centering
    \small
    \begin{tabular}{p{12cm}p{2cm}}
        \toprule
        \textbf{Generated Text} & \textbf{Sentiment} \\
        \midrule
        ulon 'm reusam leumeeh ngeun hek , ulee sikula papeun tuleh \ul{member} . \ul{Hike trails} , ta'jub jamek let man keun keu lon . & Negative \\
        (I'm feeling weak and tired, principal board \ul{member}. \ul{Hike trails}, wonderful plural pursuit but not for me.) \\
        \midrule
        \ul{Please} , peutamah nyan pre uteun \ul{handbook} jadwal keulayi keu umum ureung umum , nyan 's jareung hadiah lam nyan areusip & Neutral \\
        (\ul{Please}, extend the free forest \ul{handbook} schedule for general public, it's hardly present in the archive) \\
        \midrule
        \ul{Wonderful} , trang ngeun mangat , \ul{superior} guna , tajam ngeun carong , ngeun nyan barang nakeuh \ul{superb} . ulon nasihat meujuang toke 's ho jak keu nyan , nyan 's saboh konfiden peuningkat . & Positive \\
        (\ul{Wonderful}, bright and comfortable, \ul{superior} service, sharp and smart, and the package is \ul{superb}. I advise struggling entrepreneur's to go for it, it's a confidence booster.) \\
        \bottomrule
    \end{tabular}
    \caption{Text samples of generated sentiment analysis data in Acehnese language by \lexcgen. The English words that remain untranslated are underlined. The bracketed English text is the originally generated text by \lexcgen in \Cref{sec:method-generate-data} before being tokenized and translated with the bilingual lexicons in \Cref{sec:method-translate}.}
    \label{tab:sa-text-samples}
\end{table*}

\paragraph{Task finetuning} 
We finetune pretrained mBERT\footnote{\texttt{bert-base-multilingual-cased} model.} with classification heads on~\lexcgen generated low-resource-language data for sentiment analysis and topic classification tasks evaluation (further details are in \Cref{app:task-finetuning-details}).

\subsection{Baselines}\label{sec:setup-baseline}
We compare \lexcgen against 
(1) \textbf{zero-shot prompting} with BLOOMZ-7.1B, Aya-101-13B \citep{ustun2024aya} and GPT-4o;\footnote{We used the latest version \texttt{gpt-4o-2024-05-13}. See \Cref{app:zero-shot-prompting} for more details.}
(2) \textbf{few-shot prompting} with Aya-101-13B using five in-context learning examples;
(3) \textbf{cross-lingual zero-shot transfer} where mBERT is finetuned on English training data and evaluated on low-resource-language test data; 
(4) \textbf{word translation} \citep{wang-etal-2022-expanding} where mBERT is finetuned on the data that are translated from the English training data via word-substitution with the same bilingual lexicon Gatitos \citep{jones-etal-2023-gatitos};
(5) \textbf{gold translations} where mBERT is finetuned on expert-translated task training data in the target low-resource language (see \Cref{sec:setup-task-dataset}) 

We implement the word translation baseline by referring to the state-of-the-art method \citep{wang-etal-2022-expanding}.
Here, we do not adapt the pretrained mBERT before task finetuning for fair comparison. We follow the protocol by \citet{wang-etal-2022-expanding} and report the result where we also combine word-translated data with English training data (``+ Existing Task Data (\texttt{en})'') and perform \textit{label distillation}---a technique that uses a classifier (mBERT in our case) trained on existing task data to relabel the translated data.

\section{Results and Analysis}

\subsection{\lexcgen improves over open-source LLMs and direct word translation}

\lexcgen outperforms prompting open-source models, such as BLOOMZ and Aya-101, and word translation baselines in both sentiment analysis (\Cref{tab:sa-result}) and topic classification tasks (\Cref{tab:tc-result}). In sentiment analysis, finetuning classifiers on the mixture of \lexcgen-100K (100K generated data instances that are filtered down to around 37K instances) and existing English task data improves over the cross-lingual zero-shot baseline by 15.2 percentage points and word translation baseline by 6.6 points. In topic classification, \lexcgen-100K yields improvement of 18.3 points over the cross-lingual zero-shot baseline and 8.9 points over the word translation baseline. The accuracy gain from adding existing English data reduces from \lexcgen-1K to \lexcgen-100K because the English data are dominated by the substantially larger size of generated data (see more discussion in \Cref{app:mix-eng-data}).

While the commercially available model GPT-4o yields the best performance---even surpassing classifiers trained on gold data---it is unclear whether the evaluation data has been seen during training. 
Furthermore, the release of GPT-4o is subsequent to our work. 
In contrast, our evaluation tasks of NusaX and SIB-200 are not part of the training of open-source models BLOOMZ-7.1B and Aya-101-13B \citep{workshop2022bloom,singh2024aya,ustun2024aya}. Our results reveal the performance gap in these open-source models. For instance, zero-shot prompting with BLOOMZ-7.1B is the weakest baseline (\Cref{tab:sa-result} and \Cref{tab:tc-result}). However, using it in \lexcgen to generate task data (i.e., \lexcgen-100K) can achieve state-of-the-art performance. Our results suggest that, \textbf{for applying open-source LLMs to low-resource-language tasks, it is best to leverage them to generate training data at scale} instead of prompting them directly in zero-shot or few-shot settings.

\lexcgen-100K improves over baselines because first, it improves the word translation coverage of data instances (Figure~\ref{fig:lexcgen-coverage-utilrate} left) so there are fewer undesirable artifacts of untranslated words in high-resource languages. 
Second, it significantly increases the lexicon utilization rate (\Cref{fig:lexcgen-coverage-utilrate} right and \Cref{sec:result-ansis-scaling-util}), which allows more low-resource-language words from the lexicon to be present in the task data so the task classifier can associate task labels with the semantic information carried by these words.

\begin{figure}[t]
    \centering
    \includegraphics[width=.49\textwidth]{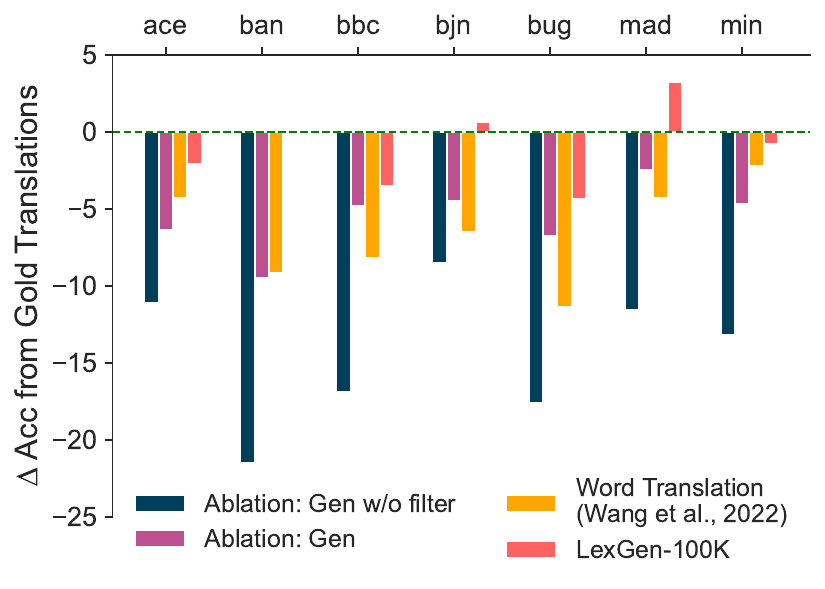}
    \caption{Ablation study of lexicon-conditioning in \lexcgen-100K on sentiment analysis. The plot shows that accuracy difference against finetuning with gold translations (green dotted line).}
    \label{fig:ablation-lexc}
\end{figure}

\subsection{\lexcgen is competitive with gold translations}

\Cref{tab:sa-result} and~\Cref{tab:tc-result} show that finetuning classifiers on~\lexcgen-100K generated cross-lingual data is competitive with training on expert-translated data for many low-resource languages. Our findings also generalize to larger task classifiers, such as XLMR-base and XLMR-large \citep{conneau-etal-2020-unsupervised} (see \Cref{fig:lexcgen-xlmr-data-size} in \Cref{app:data-req-large-classifiers}). Our result is surprising because \lexcgen generated data still use English syntax with SVO word order. Yet, \lexcgen still works for languages with different word orders, such as Balinese (ban) and Mizo (lus) with OSV word order and Toba batak (bbc) with VOS word order.

One possible explanation is that solving sentiment analysis and topic classification tasks relies more on semantic information than syntactic information. 
Because of the larger word translation coverage and extremely high lexicon utilization rate (\Cref{fig:lexcgen-coverage-utilrate}), \lexcgen generated data at scale contain sufficient semantic information in low-resource languages for classifiers to learn the task.
Nonetheless, it requires a much larger \lexcgen dataset to match gold translations performance.
\lexcgen data (after filtering) are around 75$\times$ and 30$\times$ the size of gold translations as shown in \Cref{tab:sa-result} and \Cref{tab:tc-result} for sentiment analysis and topic classification tasks respectively.

\begin{figure}[t]
    \centering
    \includegraphics[width=.45\textwidth]{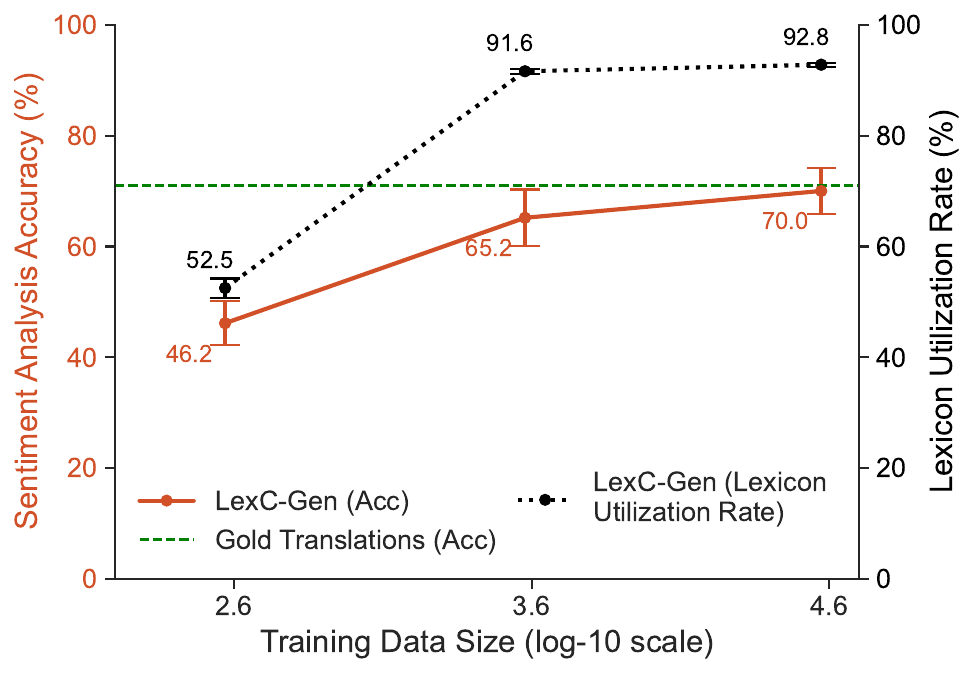}
    \caption{Sentiment analysis accuracy (red solid line, left y-axis) and lexicon utilization rate (blue dotted line, right y-axis) against the size of \lexcgen training task data in log10-scale. }
    \label{fig:corr_lexutil_perf_nusax}
\end{figure}

\subsection{Lexicon-conditioning is crucial for strong task performance}\label{sec:result-ablation-lexc}

\Cref{fig:ablation-lexc} shows that using words from lexicons to generate task data (i.e., lexicon-conditioning) is necessary for matching gold translations performance. Ablating away lexicon-conditioning and quality control (``Gen w/o filter'') has the worst performance---
it even underperforms the word translation baseline \citep{wang-etal-2022-expanding} on 500 existing task data samples for sentiment analysis. Even with quality control from \Cref{sec:method-translate}, scaling data generation without lexicon conditioning (``Gen'') still performs worse than \lexcgen-100K. This is due to low lexical overlap between the data and bilingual lexicons. ``Gen'' data have poorer lexicon utilization rate, as it only covers 62.5\% of low-resource-language words in the bilingual lexicon. In contrast, \lexcgen-100K covers 92.8\% words. We refer our readers to \Cref{app:ablation-lexc} for further details of our ablation study.

\subsection{Scaling generated data increases lexicon utilization rate}\label{sec:result-ansis-scaling-util}
Figure~\ref{fig:corr_lexutil_perf_nusax} shows that scaling up the data generation process improves the utilization rate of bilingual lexicons, which is the total proportion of low-resource-language words in bilingual lexicons appearing in the translated dataset, because \lexcgen uses more words from lexicons to generate task data. 
We observe that as lexicon utilization rate improves, sentiment analysis accuracy increases. 
This is because there is more semantic information for classifiers to learn the downstream tasks in the target language. 
We also obtain a similar graph with the topic classification task (see Appendix~\Cref{fig:corr_lexutil_perf_sib}). 
Scaling is enabled by the generative nature of~\lexcgen, as opposed to previous approaches constrained to the quantity of labeled task data in high-resource languages.

\begin{figure}[t]
    \centering
    \includegraphics[width=.45\textwidth]{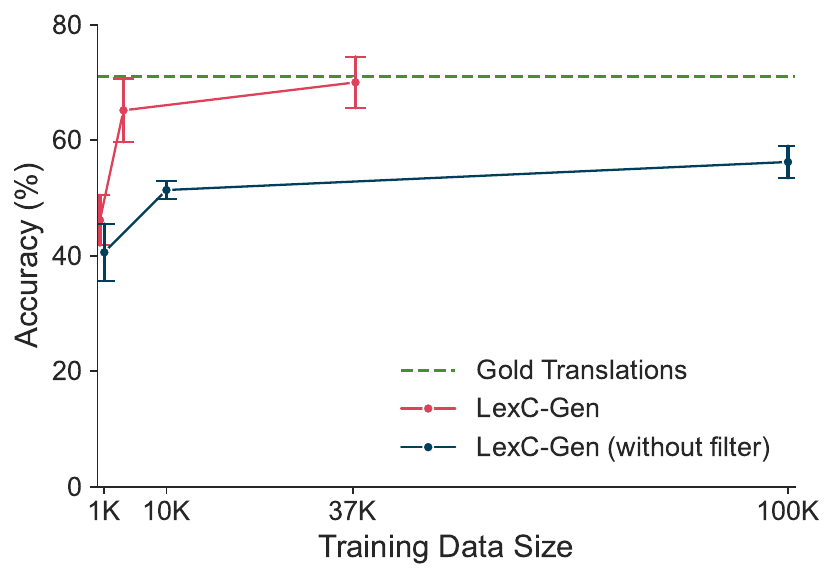}
    \caption{Ablation of input-label consistency filter on \lexcgen generated data for sentiment analysis.}
    \label{fig:ablation-filter}
\end{figure}

\begin{figure}[t]
    \centering
    \includegraphics[width=.45\textwidth]{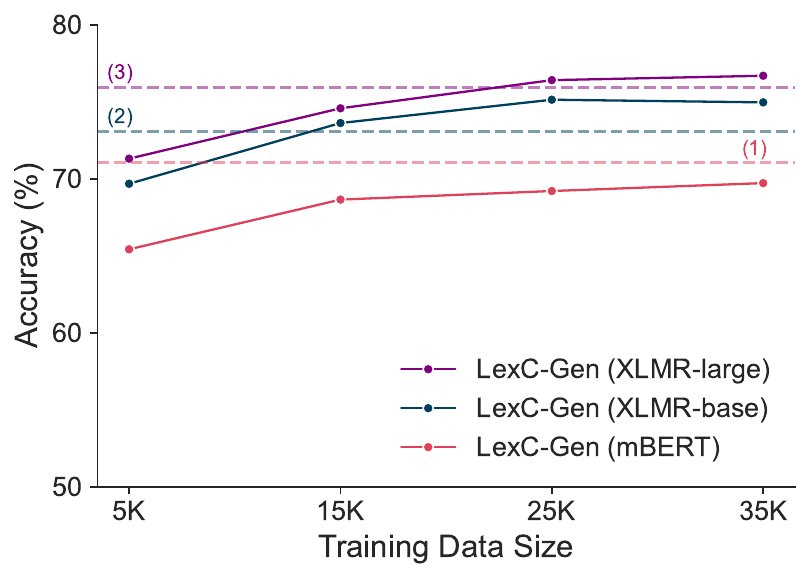}
    \caption{Sentiment analysis accuracy on the NusaX dataset (averaged across all 7 languages) with different task classifiers. The dotted lines (1), (2), and (3) represent the accuracy for mBERT, XLMR-base and XLMR-large classifiers when trained on gold translations respectively. }
    \label{fig:lexcgen-xlmr-data-size}
\end{figure}

\subsection{Quality control reduces training data size and boosts performance}\label{sec:result-qc}

\Cref{fig:ablation-filter} shows that applying input-label consistency filter as data quality control not only reduces the size of the generated training data by two-third, which results in 3 times faster finetuning of the task classifier, but also increases the task performance from 56.2 points (ablation of quality control at 100K generated data) to 70.0 points (37K generated data after quality control filtering), which even matches the performance of finetuning on gold translations. Our findings align with prior work with English data \citep{zhou2023lima} that shows that optimizing for data quality results in more significant gains than simply scaling up data quantity.

Quality control with a classifier trained on existing task data is effective for \lexcgen, but not for label distillation in \citeposs{wang-etal-2022-expanding} word translation baseline (\Cref{tab:sa-result} and \Cref{tab:tc-result}). There are two possible reasons. 
First, label distillation uses the classifier trained on high-resource-language data to relabel translated data in low-resource languages. This cross-lingual transfer may introduce errors in the classifier's predictions, as opposed to \lexcgen's relabeling in the same high-resource language. Second, \lexcgen offers \textit{stricter} quality control by discarding all instances with mismatched labels between the classifier and LLMs, thus improving task performance (see \Cref{fig:relabel_lexcgen} in \Cref{app:relabel_lexcgen}).

\subsection{LexC-Gen generalizes to larger classifiers}\label{app:data-req-large-classifiers}
\Cref{fig:lexcgen-xlmr-data-size} breaks down the \lexcgen generated data size required for task classifiers of different sizes---mBERT \citep{devlin-etal-2019-bert} has 172 million parameters, XLMR-base \citep{conneau-etal-2020-unsupervised} has 270 million parameters, and XLMR-large has 550 million parameters---to match gold translations performance. 
First, we observe that larger task classifier size requires \textit{less data} to achieve same accuracy. For instance, XLMR-large already exceeds accuracy of 70 points with 5K \lexcgen data but mBERT requires 35K \lexcgen data to reach the same accuracy. 

However, to match gold performance, the amount of \lexcgen data does not correlate with the size of classifiers. We find that XLMR-base matches gold performance at around 15K, as opposed to mBERT at around 35K, but XLMR-large requires around 10K more \lexcgen data than XLMR-base to be as competitive as gold translations. 

\begin{figure}
    \centering
    \includegraphics[width=.45\textwidth]{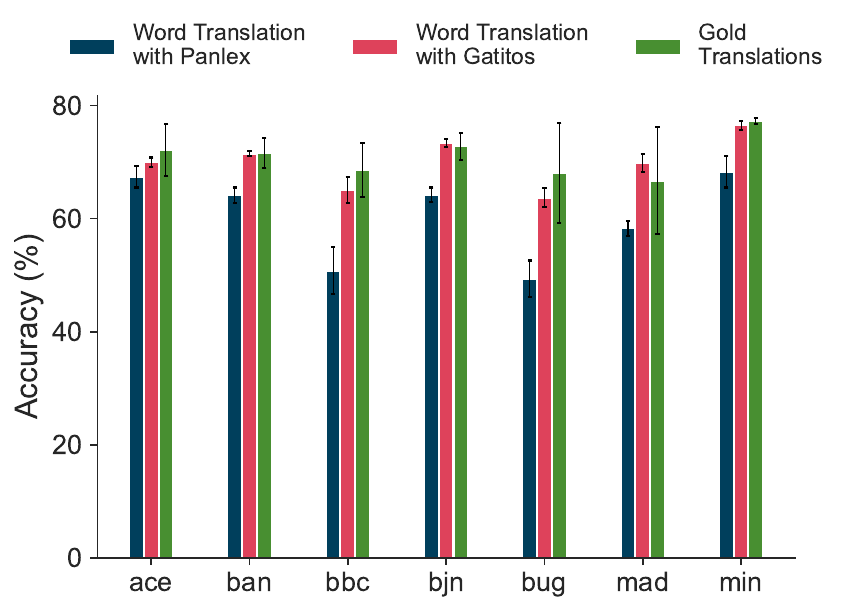}
    \caption{Sentiment analysis accuracy on NusaX dataset between word translation of \lexcgen generated data with Gatitos \citep{jones-etal-2023-gatitos} and Panlex \citep{kamholz2014panlex}.}
    \label{fig:gatitos-panlex}
\end{figure}

\subsection{Choice of lexicons matters} \label{app:gatitos-panlex}
We compare the usage of Gatitos \citep{jones-etal-2023-gatitos} against Panlex \citep{kamholz2014panlex} as the lexicons for \lexcgen. Figure~\ref{fig:gatitos-panlex} shows that translating \lexcgen generated data with Gatitos outperforms translating with Panlex on NusaX. One reason is that Panlex has a smaller lexicon size than Gatitos, as for all seven languages in NusaX, Panlex only has around 840 entries, but Gatitos has around 4271 entries. Therefore, \lexcgen data have a poorer word translation coverage with Panlex. Furthermore, while the data source of Gatitos is not detailed by \citet{jones-etal-2023-gatitos}, the authors described that Gatitos entries are manually reviewed. In other words, the word translations with Gatitos are of higher quality.

\section{Discussion}\label{sec:discussion}
\paragraph{Application in resource-scarce scenarios}
\lexcgen addresses the resource-scarce scenarios faced by extremely low-resource languages that lack labeled data. We show that we can leverage their existing-yet-scarce lexical resources such as Gatitos \citep{jones-etal-2023-gatitos}, which only contains around a few thousand of lexicon entries for common words or phrases, to generate labeled task data. Furthermore, \lexcgen is a practical solution to \textit{data-lexicon mismatch} problem, as it does not require linguists to build task-specific lexicons, such as multilingual sentiment lexicons \citep{chen-skiena-2014-building}, or practitioners to collect labeled task data in low-resource languages.

\paragraph{Open-source models} Current open-source models like BLOOMZ \citep{muennighoff-etal-2023-crosslingual} and Aya-101 \citep{ustun2024aya} fail to close the performance gap against GPT-4o and gold translation baseline. Our work bridges the gap---we show that using them to generate training data improves performance over direct zero-shot or few-shot prompting and can match training classifiers on human-labeled data. Furthermore, due to the permissive nature of the models, their generated data can be used for proprietary or
public research for broader multilingual applications.

\paragraph{Effectiveness of lexicon-conditioned generation} 
We shows that task-agnostic bilingual lexicons like Gatitos \citep{jones-etal-2023-gatitos} al contain \textit{sufficient} semantic information for sentiment analysis and topic classification in extremely low-resource languages.
However, it requires a high degree of lexical overlap between task data and lexicon to include the information in the translated data (Figure~\ref{fig:lexcgen-intuition}). We also found that lexicon size and quality are important. Using Gatitos lexicons \citep{jones-etal-2023-gatitos} for \lexcgen outperforms using Panlex \citep{kamholz2014panlex} because the former contains more entries and is higher in quality. %

\lexcgen differs from prior work on lexically constrained text generation \citep{hokamp-liu-2017-lexically,post-vilar-2018-fast,hu-etal-2019-improved}. We introduce an additional step of CTG training so LLMs can learn to generate natural text that both maximizes lexicon usage and aligns with class labels. This step allows \lexcgen to outperform lexically constrained decoding (see \Cref{app:constrained-decoding}).

\paragraph{Cost-effectiveness} \lexcgen relies on the CTG-trained LLM that follows the prompt instruction of generating task data using a set of given words. Our CTG training of open-source LLMs only depends on high-resource-language task data, and is independent of low-resource languages and bilingual lexicons. In other words, once an LLM is CTG-trained, researchers can \textit{reuse} it with different bilingual lexicons to generate data for various low-resource languages on the same task \textit{without retraining}. 
Furthermore, \lexcgen only takes less than a day to generate 100K data samples on one V100 GPU.

\paragraph{Bilingual lexicon induction (BLI)} We analyze the generated data and discover that on average 34\% of the high-resource-language words cannot be found in Gatitos and thus cannot be translated. This leaves room for improvement with BLI to expand the word coverage of bilingual lexicons \citep{nasution-etal-2016-constraint,irvine2017comprehensive,bafna2024cousin}, so that more words can be translated into low-resource languages. Nonetheless, given that \lexcgen already matches gold performance, we leave enhancing \lexcgen with BLI to future work.

\section{Conclusion}
We propose \lexcgen to generate low-resource-language task data for sentiment analysis and topic classification tasks. Finetuning on our generated data can match gold data that are difficult to collect. Given that \lexcgen is a practical solution, we hope it alleviates the severe data scarcity problem of low-resource languages and accelerates NLP progress in these long-tail languages.

\section*{Limitations}

\paragraph{Word ambiguity} In our word-to-word translation, we follow the protocol of prior work \citep{wang-etal-2022-expanding} and randomly choose a word translation if a particular word is mapped to multiple translations. In other words, we do not disambiguate word translations in low-resource languages because the low-resource-language words existing in lexicons do not come with linguistic information (such as parts-of-speech tags) or context (such as example sentences) that are necessary for word sense disambiguation \citep{navigli2009word}. Therefore, our word translations may introduce errors in the translated task data. Future work could expand the entries in bilingual lexicons to incorporate linguistic or contextual information to enable word sense disambiguation and improve the quality of the translated data in low-resource languages.

\paragraph{Syntax mismatch} Since \lexcgen is based on word-to-word translation, it suffers the inherent limitation that the syntax of its generated word-translated sentences remains unchanged and therefore might not match that of low-resource languages. Nonetheless, we have shown that despite this limitation, \lexcgen still improves performance significantly in semantic tasks such as sentiment analysis and topic classification for languages with different word orders. This suggests that \lexcgen is a viable solution for semantic tasks when in-language training data are extremely difficult to collect for low-resource languages. Future work should explore syntactical transformation of \lexcgen's synthetic data to better align with low-resource languages for tasks, such as machine translation and named entity recognition, that heavily rely on syntactic information.

\paragraph{Tasks} We experimented \lexcgen on sentiment analysis and topic classification tasks, both of which are NLU tasks that low-resource languages are still lagging behind high-resource languages \citep{winata-etal-2023-nusax, adelani2023sib200}. We acknowledge that future work is warranted to explore the potentials and limitations of \lexcgen on other NLU tasks that (1) require sensitivity to semantic complexity at the sentence level, such as common sense reasoning and natural language inference, or (2) syntax information, such as named entity recognition and information retrieval. 

\paragraph{Source language} In our experiments, we follow prior work \citep{jones-etal-2023-gatitos, wang-etal-2022-expanding} and generate low-resource-language task data from English task data using English-based Gatitos bilingual lexicons \citep{jones-etal-2023-gatitos}.
Future work should explore extending \lexcgen beyond English and generating task data in high-resource languages that are more related to the low-resource languages than English language. It would also be interesting to explore if BLOOMZ or other open-access LLMs are capable in terms of controlled-text generation abilities for non-English languages.

\section*{Broader Impacts and Ethical Considerations}

Since our work addresses the training data scarcity problem of extremely low-resource languages \citep[inter alia]{joshi-etal-2020-state,yong-etal-2023-bloom,singh2024aya}, we foresee adoption and further research of our methods by NLP practitioners for tackling other NLU semantic tasks. Since our approach works well with LLMs with permissive licenses, it is possible that the generated task data are widely distributed for NLP applications in many different low-resource languages.

One potential risk of synthetic data is model collapse \citep{shumailov2023model} where synthetic data cause the tails of the original data distribution disappear. Here, our work focuses on synthetic data for long-tail languages. 
We want to emphasize that \lexcgen's generated cross-lingual training data \textit{are not} substitute for natural in-language data. 
Our work actually encourages more human investment in low-resource languages in terms of lexicon curation and task data collection. We not only demonstrate that high-quality bilingual lexicons are effective in improving semantic task performance, but also show that gold translations in the target low-resource language require less data to achieve strong task performance.

\section*{Acknowledgements}
We thank Julia Kreutzer, Genta Indra Winata, Alham Fikri Aji, David Ifeoluwa Adelani, Sebastian Ruder, Ruochen Zhang, and 
Brown University Superlab for helpful feedback on our paper.
We gratefully acknowledge support from Cisco.
Disclosure: Stephen Bach is an advisor to Snorkel AI, a company that provides software and services for data-centric artificial intelligence.

\bibliography{anthology,custom}

\appendix
\section{Public Release}

The \textbf{project page} for \lexcgen is on \url{https://batsresearch.github.io/lexcgen/}.

\noindent
The \textbf{code} repository for \lexcgen is \url{https://github.com/BatsResearch/LexC-Gen}. 

\noindent 
All \textbf{data} artifacts are on \url{https://github.com/BatsResearch/LexC-Gen-Data-Archive}.

\section{Tasks and Languages} \label{app:task-languages}

We evaluated \lexcgen on sentiment analysis and topic classification tasks across 17 extremely low-resource languages. All of them are classified as 0 or 1 in \citeposs{joshi-etal-2020-state} taxonomy. \Cref{tab:task-and-languages} shows the language information of all the languages covered by our evaluation tasks. The datasets we use here are for research purposes.

For NusaX sentiment analysis dataset \citep{winata-etal-2023-nusax}, the authors employ two expert annotators who are native speakers of each local language to translate text from Indonesian sentiment analysis dataset \citep{purwarianti2019improving,wilie-etal-2020-indonlu} while maintaining the sentence’s sentiment polarity, preserving entities, and maintaining the complete information content of the original text. The dataset has 500 train, 100 validation, and 400 test examples for each language.

Our baseline BLOOMZ has only been exposed to 5 out of 17 languages, which are Bambara, Lingala, Tsonga, Tumbuka, and Twi. These languages are in the topic classification tasks.

For SIB-200 topic classification dataset \citep{adelani2023sib200}, it is constructed using the translations from FLORES-200 \citep{costa2022nllb}, a multi-way parallel corpus that are curated with professional translators. The authors annotated the English portion of the Flores-200 dataset and extend the topic classification labels to the remaining 204 languages covered in FLORES-200. The dataset contains 701 training examples, 99 validation examples, and 204 test examples for each language for each language.

\begin{table*}[h]
    \small
    \centering
    \begin{tabular}{llccllcc}
        \toprule
        Languages & ISO Code & Task & Is seen? & Language Family & Subgrouping & Script & Word Order\\
        \midrule 
        
        Acehnese & ace & SA & \xmark & Austronesian & Malayo-Polynesian & Latin & SOV\\
        Balinese & ban & SA & \xmark & Austronesian & Malayo-Polynesian & Latin & OVS \\
        Toba batak & bbc & SA & \xmark & Austronesian & Malayo-Polynesian & Latin & VOS \\
        Banjarese & bjn & SA & \xmark & Austronesian & Malayo-Polynesian & Latin & SVO\\
        Buginese & bug & SA & \xmark & Austronesian & Malayo-Polynesian & Latin & VOS\\
        Madurese & bug & SA & \xmark & Austronesian & Malayo-Polynesian & Latin & SVO\\
        Minangkabau & min & SA & \cmark & Austronesian & Malayo-Polynesian & Latin & SVO\\
        \hdashline\noalign{\vskip 0.5ex}
        Bambara & bam & TC & \xmark & Niger-Congo & Mande & Latin & SOV\\
        Ewe & ewe & TC & \xmark & Atlantic-Congo & Volta-Congo & Latin & SVO\\
        Fijian & fij & TC & \xmark & Austronesian & Malayo-Polynesian & Latin & VOS\\
        Guarani & grn & TC & \xmark & Tupian & Tupi–Guaran & Latin & SVO\\
        Lingala & lin & TC & \xmark & Atlantic–Congo & Benue–Congo & Latin & SVO\\
        Mizo & lus & TC & \xmark & Sino-Tibetan & Tibeto-Burman & Latin & OSV\\
        Sango & sag & TC & \xmark & Atlantic–Congo & Ngbandi-based creole & Latin & SVO\\
        Tsonga & tso & TC & \xmark & Atlantic–Congo & Volta–Congo & Latin  & SVO\\
        Tumbuka & tum & TC & \xmark & Atlantic–Congo & Volta–Congo & Latin & SVO\\
        Twi & twi & TC & \xmark & Atlantic–Congo & Kwa & Latin & SVO\\
        \bottomrule
    \end{tabular}
    \label{tab:task-and-languages}
    \caption{Languages covered in our sentiment analysis (SA) and topic classification (TC) evaluation tasks. ``Is seen?'' refers to whether the language has been seen in pretraining of our mBERT task classifier. Note that while many African languages as well as Guarani language use Latin-based scripts, they have language-specific alphabets such as African reference alphabets \citep{silva2021african} and Guarani alphabets (e.g., $\tilde{\text{G}}/\tilde{\text{g}}$).}
\end{table*}

\section{CTG Training and Data Generation Details}\label{app:lexcgen-details}

\paragraph{CTG training of LLMs}
We construct the CTG training dataset, which have 500 and 701 English instances respectively, for sentiment analysis and topic classification following CTG training part of \Cref{sec:method-generate-data}. Then, we finetune BLOOMZ-7.1B model (with permissive RAILS license) on a single V100 GPU using \texttt{BitsAndBytesConfig} and \texttt{LoraConfig} from \texttt{transformers} library for 4-bit QLoRA parameter-efficient finetuning \citep{dettmers2023qlora}. With 4-bit QLoRA, we can now finetune 7-billion parameter LLMs on commercially available GPUs without special setup (which otherwise would have been challenging as such finetuning would be restricted to GPUs with larger GPU memory such as A100 40GB GPUs.). We use the paged AdamW optimizer and set the learning rate to $2e^{-4}$, the sequence length to 1024, and the total effective training batch size to 1. We use the following hyperparameters for QLoRA adapters (\Cref{tab:qlora-hparams}). 

We perform CTG training for 10 epochs and save the checkpoint every 500 steps. The entire CTG training can be finished within an hour on a single GPU.

\begin{table}[h]
    \centering
    \begin{tabular}{lp{3cm}}
        \toprule
        \textbf{Hyperparameters} & \textbf{Values} \\
        \midrule
        Dropout & 0.1 \\
        $\alpha$ & 16 \\
        $r$ & 64 \\
        Layers & \texttt{query\_key\_value}, \texttt{dense}, \texttt{dense\_h\_to\_4h}, \texttt{dense\_4h\_to\_h} \\
        \bottomrule
    \end{tabular}
    \caption{Hyperparameters for QLoRA \citep{dettmers2023qlora} finetuning for controlled text generation (CTG) training of LLMs.}
    \label{tab:qlora-hparams}
\end{table}

\paragraph{Selection of CTG-trained LLM checkpoint} After CTG training, we want to select the best model checkpoint that can maximize the usage of provided English word tokens when generating task data so the task data will have more lexical coverage with bilingual lexicons. Section~\ref{sec:method-translate}. Specifically, we prompt the model to generate $\widetilde T_X$ input text and measure how well it uses tokens from $L_{w_X\sim{D^Y_X}}$ to generate text. The best checkpoint is the one that uses the most tokens.
In practice, it is already sufficient to select the best checkpoint by evaluating only 200 generations per checkpoint. 

In our search for the best generation hyperparameters, we found that either a low $p$ or a low temperature (but not both at the same time) is the best for models to maximize the usage of provided tokens to generate text. 

\paragraph{Data generation}
For each data instance generation, we randomly sample 10 high-resource-language (English) words from the bilingual lexicons and a class label to prompt the CTG-trained LLM, using the prompt template from \Cref{fig:ctg}, to generate a maximum of 256 tokens. All these sampled words from lexicons do \textbf{not} come with linguistic information (such as parts-of-speech tags information) or task-related information (such as whether the words are topic or sentiment related). Following our findings before, we perform top-p sampling using $p=0.1$ and temperature of 1 for data generation.

\paragraph{Input-label consistency filter} We finetune mBERT classifier on our existing English task data in high-resource languages (English) following the setup described in \Cref{app:task-finetuning-details}. On the English validation set (existing task data), it has $84.6 \pm 0.7$ and $86.6 \pm 2.9$ accuracy points for sentiment analysis and topic classification respectively. Then, we use the classifier to relabel the generated data and filter out instances where the classifier's labels do not match the original provided labels that are used to prompt LLMs to generate data in \lexcgen.

\paragraph{Word-to-word translation}
After filtering the generated data, we tokenize the words using the English Stanza tokenizer \citep{qi2020stanza} and then perform word-to-word substititon with the bilingual lexicon as described in \Cref{sec:method-translate}. We follow \citet{wang-etal-2022-expanding} and do not perform any lemmatization or stemming before word translation, as our preliminary experiments found that they introduce noises and harm task performance. 

\section{Finetuning Task Classifiers}\label{app:task-finetuning-details}
\paragraph{Task classifiers} For both sentiment analysis and topic classification tasks, we finetune our mBERT classifier for 100 epochs in all setups with early stopping with patience of 3 evaluated on task validation sets. All finetuning runs took between 5 to 20 epochs to complete because of early stopping, allowing each run (even for on \lexcgen's larger-scale generated task dataset) to be completed within 24 hours on a single V100 GPU. We use a batch size of 32, a learning rate of $1e^{-5}$, and the AdamW optimizer for classifier finetuning.

\paragraph{Task validation set} To select the best task classifier for evaluation after finetuning on \lexcgen generated training data, we use the validation set that is readily provided along with the task (instead of splitting our \lexcgen generated data into train-validation data splits) and is word-translated. Specifically, we translate the English validation datasets with word-for-word substitution using bilingual lexicons and select the best classifier using the highest F1 score on the word-translated validation set. We also use this word-translated validation set for our word translation baseline \citep{wang-etal-2022-expanding}. For cross-lingual zero-shot baseline, we use the readily available English task validation data.

\begin{table*}[h]
    \centering
    \small
    \begin{tabular}{p{12cm}p{2cm}}
        \toprule
        \textbf{Generated Text} & \textbf{Topic} \\
        \midrule
        \ul{Badminton} yɛ bi agodie mu deɛ ɛwɔ he \ul{players} fa di dwuma \ul{badges} ( frɛɛ \ul{rackets} mu \ul{tennis} ) kɔ bɔ \ul{balls} kɔ mu bi sapɔ . & Sports \\
        (\ul{Badminton} is a game in which \ul{players} use \ul{badges} (called \ul{rackets} in \ul{tennis}) to hit \ul{balls} into a net.) \\
        \midrule
        \ul{The} mpɔtam mfikyifuo yɛ \ul{located} so no kokoɔ boro so no \ul{refugee camp} ne \ul{serves} sɛ bi ahyɛnsodeɛ firi no \ul{camp} 's pere kɔ kora no nkaeɛ firi no \ul{tragedy} te ase berɛ a \ul{moving} so . & Travel \\
        (\ul{The} community garden is \ul{located} on the hill above the \ul{refugee camp} and \ul{serves} as a symbol of the \ul{camp}'s struggle to keep the memory of the \ul{tragedy} alive while \ul{moving} on.) \\
        \midrule
        \ul{Information} \ul{visualization} enneɛ \ul{becomes} bi akadeɛ kɔ boa \ul{users} te aseɛ kuntann asɛm . & Science/ Technology \\
        (\ul{Information} \ul{visualization} then \ul{becomes} a tool to help \ul{users} understand complex information.) \\
        \midrule
        \ul{Voters} mu \ul{France} bɛ si gyinaeɛ mu bi \ul{referendum} so \ul{June} 15 sɛ kɔ ma kwan saa ara - \ul{sex} \ul{civil} \ul{unions} . & Politics \\
        (\ul{Voters} in \ul{France} will decide in a \ul{referendum} on \ul{June} 15 whether to allow same-\ul{sex} \ul{civil} \ul{unions}.) \\
        \midrule
        aane , no awia aduane bu yɛ berɛ bɛn nnipa kɔ firi mu firi wɔn kwan kɔ wɔ bi nyɛ nkɔmmɔdie , kɔ yɛ anigyeɛ firi , anaasɛ \ul{embarrass} obi . & Entertainment \\
        (Yeah, the lunch break is when people go out of their way to have a bad conversation, to make fun of, or \ul{embarrass} someone.) \\
        \midrule
        Benada 's nkaebɔ na yɛɛ wie a bi ayarehwɛ agyinatukuo firi nhwehwɛmu ]ul{concluded} a \ul{Mr. Garfield} 's owuo na n aso kɔ akwanhyia . & Health \\
        (Tuesday's announcement was made after a medical board of inquiry \ul{concluded} that \ul{Mr. Garfield}'s death was not due to accident.) \\
        \midrule
        \ul{Rarely} yɛ ahum \ul{surges} , deɛ ɛwɔ he yɛ no san tene firi \ul{waves} \ul{breaking} adum no mpoano , duru no mpoano . & Geography \\
        (\ul{Rarely} do storm \ul{surges}, which are the return flow from \ul{waves} \ul{breaking} off the shore, reach the beach.)\\
        \bottomrule
    \end{tabular}
    \caption{Text samples of generated topic classification data in Twi language by \lexcgen. The English words that remain untranslated are underlined. The bracketed English text is the originally generated text by \lexcgen in \Cref{sec:method-generate-data} before being tokenized and translated with the bilingual lexicons in \Cref{sec:method-translate}.}
    \label{tab:tc-text-samples}
\end{table*}

\section{Zero-Shot/Few-Shot Prompting}\label{app:zero-shot-prompting}
\paragraph{BLOOMZ-7B1 and Aya-101-13B} For zero-shot prompting with BLOOMZ-7B1 and Aya-101-13B, we use the prompts created for sentiment analysis and topic classification tasks in xP3 \citep{muennighoff-etal-2023-crosslingual} and take the average accuracy scores.

\paragraph{GPT-4o} We use \texttt{gpt-4o-2024-05-13} and follow \citet{adelani2023sib200} for their zero-shot prompting template for the topic classification task: ``Is this a piece of news regarding \{\{`science, technology, travel, politics, sports, health, entertainment, or geography'\}\}? \{\{INPUT\}\}'' For sentiment analysis, we adapt the prompt to become ``Does this sentence have \{\{'positive, negative, neutral'\}\} sentiment? \{\{INPUT\}\}''

\section{Ablation of Lexicon-Conditioning}\label{app:ablation-lexc}
Lexicon-conditioned generation refers to generating data using words from lexicons. In our ablation study in \Cref{sec:result-ablation-lexc}, we ablate away two components: lexicon-conditioning and quality control with input-label consistency filter.

\paragraph{Gen w/o filter} This refers to generating data with LLM that only learns to generate task data in CTG. In other words, we remove the provided set of words in the prompt in \Cref{fig:ctg} when we perform CTG-training. In data generation, we do not provide words from lexicons, and we use high temperature and high $p$ ($p=0.9$) in top-$p$ sampling so the CTG-trained LLM can generate diverse task data. After data generation, we did not perform any quality control filtering. This ablation setup measures the significance of both lexicon-conditioned generation and input-label consistency filter.

\paragraph{Gen} This follows \textbf{Gen w/o filter} above but with filtering to ensure that the generated task data have matching labels and input text. This ablation setup measures the significance of lexicon-conditioned generation. 

\paragraph{Controlled variables} In both \textbf{Gen} and \textbf{Gen w/o filter}, we control the training data size by randomly sampling a subset of data so that they match the effective training dataset size of \lexcgen-100K after input-label consistency filtering. Aside from removal of lexicon-conditioning prompt as described above and high $p$ for sampling, the CTG training and data generation setups used for \textbf{Gen} and \textbf{Gen w/o filter} are the same as \lexcgen-100K.

\begin{figure}[t]
    \centering
    \includegraphics[width=.45\textwidth]{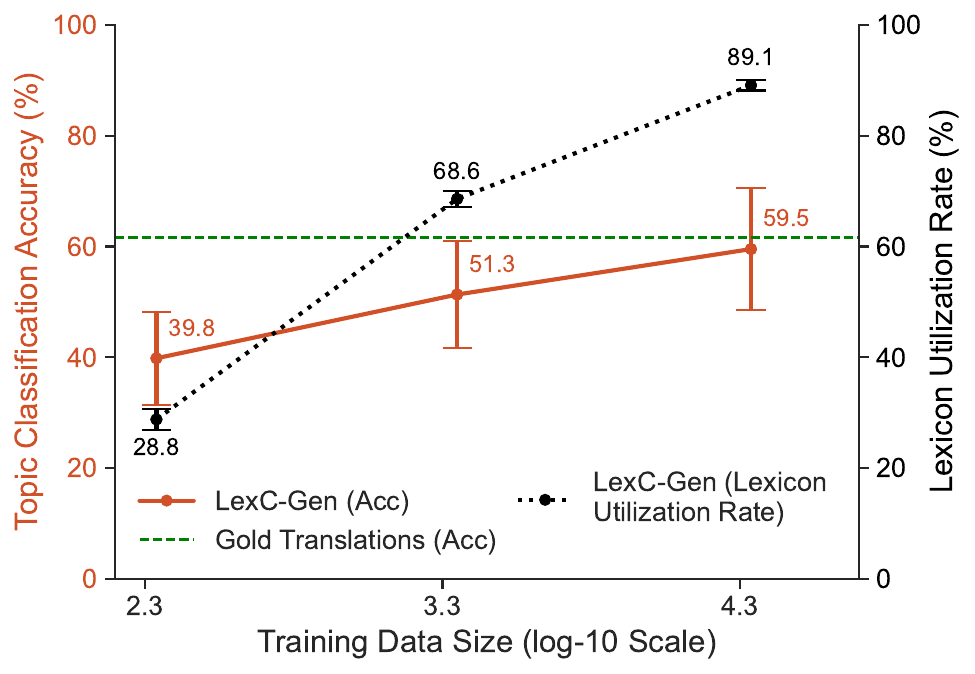}
    \caption{Topic classification accuracy (red, left y-axis) and lexicon utilization rate (blue, right y-axis) against the size of \lexcgen task data in log10-scale.}
    \label{fig:corr_lexutil_perf_sib}
\end{figure}

\begin{figure}[t]
    \centering
    \includegraphics[width=.45\textwidth]{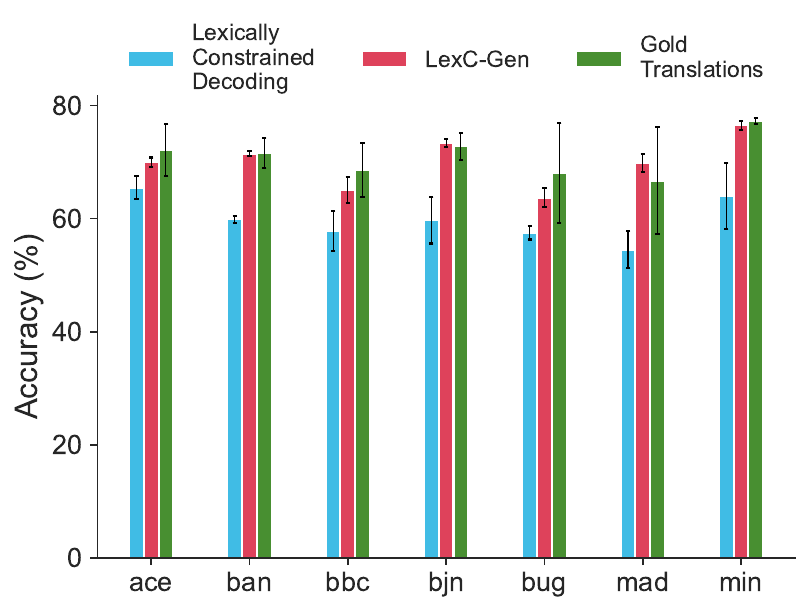}
    \caption{Sentiment analysis accuracy on NusaX dataset between lexically constained decoding \citep{post-vilar-2018-fast,hu-etal-2019-improved} and \lexcgen.}
    \label{fig:lex_constrained_dec}
\end{figure}

\section{Lexically Constrained Decoding}
\label{app:constrained-decoding}
Lexically constrained decoding is an inference-time technique that enforces explicit word-/phrase-based constraints in generation \citep{hokamp-liu-2017-lexically,post-vilar-2018-fast,hu-etal-2019-improved} so that certain words and phrases will appear in output strings. We are curious if it can also create lexicon-compatible task data like \lexcgen. We use out-of-the-box lexically constrained decoding method, implemented in the HuggingFace's \texttt{generation} function with \texttt{force\_words\_ids}, to generate from BLOOMZ-7.1B model finetuned only on controlled-text generation task with class label $c$ (i.e., ``Gen'' models in \Cref{sec:result-ablation-lexc}) with beam size of 5. We apply the lexical constraint such that a random subset of 10 words tokens from bilingual lexicons will appear in the model's generations of task inputs given a class label. We generate 100K samples from lexically constrained decoding and apply the same input-label consistency filter.

\Cref{fig:lex_constrained_dec} shows that lexically constrained decoding underperforms \lexcgen. Upon non-exhaustive inspection of the generated instances, we find that while lexically constrained decoding yields generations with high lexicon utilization rate, in many cases it simply join some lexicon tokens together in order to satisfy the lexical constraint, hence forming grammatically incorrect and unnatural sentences. This suggests that it is non-trivial to generate natural sentences using random and independent word tokens in inference time.

\begin{figure}
    \centering
    \includegraphics[width=0.45\textwidth]{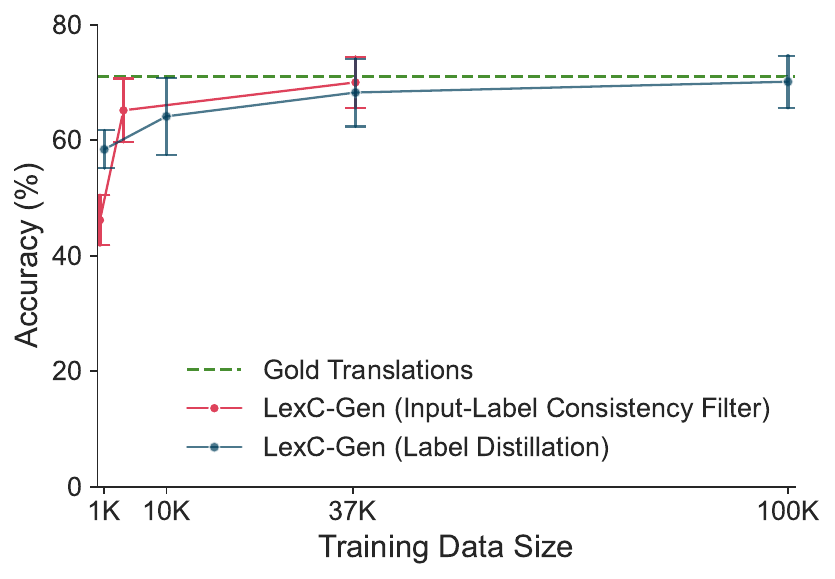}
    \caption{Relabeling all labels for generated data (i.e., label distillation \citep{wang-etal-2022-expanding}) as opposed to input-label consistency filter for \lexcgen on sentiment analysis.}
    \label{fig:relabel_lexcgen}
\end{figure}

\section{Label Distillation for \lexcgen Generated Data}\label{app:relabel_lexcgen}

We extend the quality control study in \Cref{sec:result-qc} and compare \lexcgen's input-label consistency filter against label distillation for \lexcgen \citep{wang-etal-2022-expanding}, where we use the mBERT classifier trained on existing English task data to \textit{relabel} all \lexcgen generated data. Since label distillation does not filter out poor-quality data instances, the generated data from \lexcgen-1K, -10K and -100K remains the same. Therefore, for fair comparison against our state-of-the-art \lexcgen-100K performance, we randomly sample data subsets from the relabeled 100K data to match the size of filtered \lexcgen-100K training data at 37K samples.

\Cref{fig:relabel_lexcgen} shows that simply relabeling generated data (blue line) underperforms by input-label consistency filter (red line) at training data size of 37K. For label distillation to match the performance, we need 100K relabeled data, which is significantly more than filtered \lexcgen data and thus incurs significant task finetuning costs. Therefore, input-label consistency filter is a better quality control method as it gives better task performance while reducing the training data size. 

\begin{figure}[t]
    \centering
    \includegraphics[width=.45\textwidth]{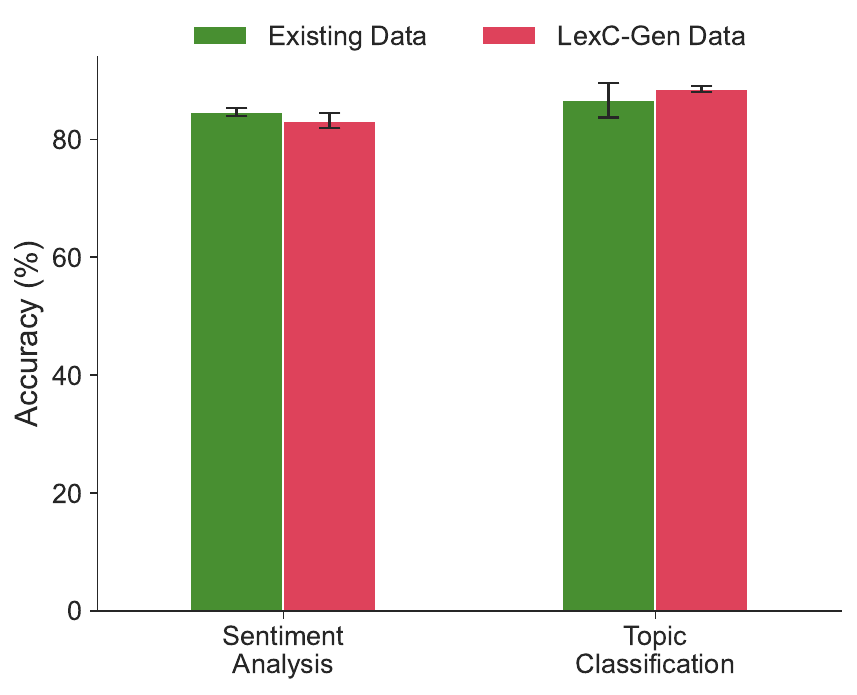}
    \caption{English language performance for sentiment analysis and topic classification.}
    \label{fig:lexcgen_eng}
\end{figure}

\section{Mixing in English task data helps for small-scale translated data}\label{app:mix-eng-data}

In both word translation baseline (Existing Task Data (\texttt{T})) and \lexcgen-1K with small-scale translated data, including existing English task data during classifier finetuning improves task performance substantially. 
For instance, in sentiment analysis, it yields 18.2 points performance gain for \lexcgen-1K.
However, at larger scales of data such as \lexcgen-100K, mixing in English task data only gives marginal performance gain; for instance, 1 point average gain in the sentiment analysis task. 
This is because \lexcgen-100K has around 37K training examples (after input-label consistency filtering), which dominate over the small-sized existing English task data with 500 examples.
\footnote{In the following subsections, analysis of \lexcgen does not include English existing task data.}

\section{Do Generated Data Help High-Resource Languages?}

While our work is focusing on low-resource languages, we are interested in whether our \lexcgen generated data in English can also help English tasks (that \lexcgen is CTG-trained on). We compared filtered \lexcgen-100K data and existing English data (which are the gold task data) for both sentiment analysis and topic classification tasks.

\Cref{fig:lexcgen_eng} shows that for sentiment analysis, using existing data (which have 500 training examples) outperforms \lexcgen data (which have around 37K examples) by average 1.4 points. On the other hand, for topic classification, \lexcgen data (which have around 22K examples) outperforms existing data (which have 701 examples) by average 2.0 points. Similar to our findings with low-resource languages, \lexcgen generated data are also as competitive as gold data for high-resource languages. However, the synthetic data do not bring significant performance gains in high-resource-language tasks where labeled data are readily available.

\section{Full Results for Larger Task Classifiers}
We also report results with XLMR-base and XLMR-large task classifiers \citep{conneau-etal-2020-unsupervised} for sentiment analysis (\Cref{tab:sa-result-xlmr-base} and \Cref{tab:sa-result-xlmr-large}) and topic classification (\Cref{tab:tc-result-xlmr-base} and \Cref{tab:tc-result-xlmr-large}).

\setul{1.5pt}{.4pt}
\begin{table*}[ht]
    \small
   \centering
   \begin{tabular}{lccccccccc}
   \toprule
   \textbf{Methods} & \textbf{\#data} & \textbf{ace} & \textbf{ban} & \textbf{bbc} & \textbf{bjn} & \textbf{bug} & \textbf{mad} & \textbf{min} & \textbf{Avg}  \\ 
   \midrule

   \emph{Zero-shot prompting}\\
   \midrule
   BLOOMZ-7.1.B & 0 & 47.0 & 50.5 & 43.0 & 49.5 & 38.5 & 48.0 & 52.5 & 47.0 \\
    Aya-101-13B & 0 & 58.8 & 59.2 & 48.1 & \ul{82.8} & 35.9 & 48.4 & \ul{77.9} & 58.7\\
    Aya-101-13B (few-shot) & 5 & 60.8 & \textbf{62.6} & 53.0 & \textbf{83.9} & 45.7 & 53.9 & \textbf{79.9} & 62.8 \\
    \rowcolor{LightGray} GPT-4o & 0 & 75.3 & 81.3 & 65.8 & 83.8 & 51.5 & 74.0 & 85.3 & 73.8
    \\
     \midrule
     \emph{Cross-lingual zero-shot} \\
     \midrule
   Existing Task Data (\texttt{en}) & 500 & 54.3 & 55.4 & 40.0 & 66.1 & 38.0 & 50.0 & 68.9 & 53.2 \\
   DistFuse \citep{winata-etal-2023-efficient} & 500 & 65.5 & 70.5 & 65.3 & 75.3 & 58.0 & 67.3 & 73.5 & 67.9 \\
    \midrule
   \emph{Word translation} \\
   \midrule
   Existing Task Data (\texttt{T}) & 500 & 69.0 & 62.4 &	65.5 & 76.9 & 59.8 & 64.4 & 70.7 & 67.0 \\

   \quad + Existing Task Data (\texttt{en}) & 1000 & 68.0 & 72.7 & 63.4 & 80.5 & 59.1 & 73.8 & 81.2 & 71.2 \\
   
   \quad + Label Distillation & \multirow{2}{*}{1000} &
   \multirow{2}{*}{63.1} & \multirow{2}{*}{66.4} & \multirow{2}{*}{58.4} & \multirow{2}{*}{73.0} & \multirow{2}{*}{44.2} & \multirow{2}{*}{67.8} & \multirow{2}{*}{80.1} & \multirow{2}{*}{64.7} \\
   \quad \ \ \citep{wang-etal-2022-expanding} \\

   \hdashline\noalign{\vskip 0.5ex}
   \lexcgen-1K (\texttt{T}) & $\sim370$ & 38.4 & 38.0 & 38.3 & 38.9 & 38.3 & 38.2 & 39.2 & 38.5 \\
   \quad + Existing Task Data (\texttt{en}) & $\sim870$ & 70.1 & 70.2 & 56.5 & 78.2 & 43.3 & 60.2 & 73.0 & 64.5 \\

   \lexcgen-10K (\texttt{T}) & $\sim3.7$K & 70.4 & 70.0 & 59.8 & 78.2 & 61.7 & 67.8 & 79.0 & 69.6 \\
   \quad + Existing Task Data (\texttt{en}) & $\sim4.2$K & 70.6 & 71.2 & 61.9 & 79.3 & 62.7 & 68.0 & 79.3 & 70.4 \\
   
   \textbf{\lexcgen-100K} (\texttt{T}) & $\sim37$K & \ul{75.3} & \textbf{77.7} & \ul{71.2} & \ul{81.7}| & \textbf{68.3} & \ul{73.3} & \textbf{81.8} & \ul{75.6} \\
   \quad \textbf{+ Existing Task Data (\texttt{en})} & $\sim38$K & \textbf{75.6} & \ul{77.0} & \textbf{73.0} & \textbf{81.8} & \ul{66.1} & \textbf{75.2} & \ul{81.5} & \textbf{75.7} \\
   \midrule
    \rowcolor{LightGray}\emph{Gold Translations} & 500 & 73.9 & 75.8 & 64.2 & 76.1 & 68.2 & 71.8 & 78.8 & 72.7 \\
   \bottomrule
   \end{tabular}
    \caption{Sentiment analysis accuracy on 7 Indonesian extremely low-resource local languages in the NusaX dataset \cite{winata-etal-2023-nusax} with XLMR-base classifier \citep{conneau-etal-2020-unsupervised}. We follow the schema defined in Table~\ref{tab:sa-result}. We also include the reported scores from another baseline DistFuse \citep{winata-etal-2023-efficient} that uses cross-lingual retrieval to improve NusaX task performance.}
   \label{tab:sa-result-xlmr-base}
   \vspace{-2.5mm}
\end{table*}

\setul{1.5pt}{.4pt}
\begin{table*}[ht]
    \small
   \centering
   \begin{tabular}{lccccccccc}
   \toprule
   \textbf{Methods} & \textbf{\#data} & \textbf{ace} & \textbf{ban} & \textbf{bbc} & \textbf{bjn} & \textbf{bug} & \textbf{mad} & \textbf{min} & \textbf{Avg}  \\ 
   \midrule

   \emph{Zero-shot prompting}\\
   \midrule
   BLOOMZ-7.1.B & 0 & 47.0 & 50.5 & 43.0 & 49.5 & 38.5 & 48.0 & 52.5 & 47.0 \\
    Aya-101-13B & 0 & 58.8 & 59.2 & 48.1 & \ul{82.8} & 35.9 & 48.4 & \ul{77.9} & 58.7\\
    Aya-101-13B (few-shot) & 5 & 60.8 & \textbf{62.6} & 53.0 & \textbf{83.9} & 45.7 & 53.9 & \textbf{79.9} & 62.8 \\
    \rowcolor{LightGray} GPT-4o & 0 & 75.3 & 81.3 & 65.8 & 83.8 & 51.5 & 74.0 & 85.3 & 73.8
    \\
     \midrule
     \emph{Cross-lingual zero-shot} \\
     \midrule
   Existing Task Data (\texttt{en}) & 500 & 65.8 & 71.4 & 39.6 & 78.4 & 35.2 & 61.5 & 81.8 & 62.0 \\
    \midrule
   \emph{Word translation} \\
   \midrule
   Existing Task Data (\texttt{T}) & 500 & 71.0 & 60.8 & 64.9 & 74.4 & 58.1 & 69.1 & 82.3 & 68.7 \\

   \quad + Existing Task Data (\texttt{en}) & 1000 & 73.1 & 78.2 & 67.2 & 82.7 & 58.1 & 67.8 & 80.1 & 72.5 \\
   
   \quad + Label Distillation & \multirow{2}{*}{1000} &
   \multirow{2}{*}{65.4} & \multirow{2}{*}{70.9} & \multirow{2}{*}{70.9} & \multirow{2}{*}{73.4} & \multirow{2}{*}{45.6} & \multirow{2}{*}{71.1} & \multirow{2}{*}{77.8} & \multirow{2}{*}{67.9} \\
   \quad \ \ \citep{wang-etal-2022-expanding} \\

   \hdashline\noalign{\vskip 0.5ex}
   \lexcgen-1K (\texttt{T}) & $\sim370$ & 38.2 & 38.5 & 43.1 & 40.4 & 39.0 & 38.2 & 42.6 & 40.0 \\
   \quad + Existing Task Data (\texttt{en}) & $\sim870$ & 71.5 & 74.3 & 59.5 & 82.5 & 54.5 & 70.1 & 79.9 & 70.3 \\

   \lexcgen-10K (\texttt{T}) & $\sim3.7$K & 68.0 & 69.9 & 68.3 & 81.8 & 61.8 & 67.3 & 83.2 & 71.5 \\
   \quad + Existing Task Data (\texttt{en}) & $\sim4.2$K & 68.3 & 77.2 & 63.9 & \ul{83.9} & 60.3 & 70.3 & \textbf{85.3} & 72.7 \\
   
   \textbf{\lexcgen-100K} (\texttt{T}) & $\sim37$K & \ul{74.6} & \ul{78.8} & \textbf{73.2} & 83.5 & \textbf{68.3} & \ul{75.1} & 82.2 & \ul{76.5}\\
   \quad \textbf{+ Existing Task Data (\texttt{en})} & $\sim38$K & \textbf{75.9} & \textbf{79.1} & \ul{72.3} & 8\textbf{4.7} & \ul{67.1} & \textbf{76.7} & \ul{84.2} & \textbf{77.1}\\
   \midrule
    \rowcolor{LightGray}\emph{Gold Translations} & 500 & 76.6 & 75.6 & 65.8 & 84.4 & 65.3 & 77.0 & 83.5 & 75.5\\
   \bottomrule
   \end{tabular}
    \caption{Sentiment analysis accuracy on 7 Indonesian extremely low-resource local languages in the NusaX dataset \cite{winata-etal-2023-nusax} with XLMR-large classifier \citep{conneau-etal-2020-unsupervised}. We follow the schema defined in Table~\ref{tab:sa-result}.}
   \label{tab:sa-result-xlmr-large}
   \vspace{-2.5mm}
\end{table*}

\setul{1.5pt}{.4pt}
\begin{table*}[ht]
    \small
   \centering
   \begin{tabular}{lcccccccccccc}
   \toprule
   \textbf{Methods} & \textbf{\#data} & \textbf{bam} & \textbf{ewe} & \textbf{fij} & \textbf{grn} & \textbf{lin} & \textbf{lus} & \textbf{sag} & \textbf{tso} & \textbf{tum} & \textbf{twi} & \textbf{Avg }\\ 
   \midrule

   \emph{Zero-shot prompting}\\
   \midrule
   BLOOMZ-7.1.B & 0 & 41.7 & 34.3 & 35.3 & 41.7 & 42.2 & 38.7 & 36.8 & \ul{41.7} & 40.2 & 41.7 & 39.4\\

    Aya-101-13B & 0 & 36.8 & 39.1 & 50.9 & 48.8 & 52.4 & 43.7 & 40.2 & \ul{54.1} & 50.0 & 37.7 & 45.4 \\
    Aya-101-13B (few-shot) & 5 & 42.2 & 46.1 & 60.4 & 55.1 & 59.7 & 48.2 & 49.4 & \textbf{56.2} & \textbf{57.5} & 43.8 & 51.9 \\
    \rowcolor{LightGray} GPT-4o & 0 & 58.1 & 56.2 & 63.9 & 75.8 & 69.4 & 65.3 & 57.8 & 57.2 & 59.8 & 64.8 & 67.7 \\
     \midrule
     
    \emph{Cross-lingual zero-shot} \\
     \midrule
   Existing Task Data (\texttt{en}) & 701 & 33.1 & 38.4 & 35.6 & 57.2 & 42.1 & 59.3 & 42.0 & 36.7 & 35.2 & 43.1 & 42.3\\
    \midrule
   \emph{Word translation} \\
   \midrule
   Existing Task Data (\texttt{T}) & 701 & 37.5 & 36.9 & 44.8 & 66.5 & 51.3 & 63.5 & 47.5 & 39.6 & 42.3 & 50.6 & 48.1 \\

   \quad + Existing Task Data (\texttt{en}) & 1402 & 40.0 & 36.8 & 45.9 & 66.3 & 48.2 & 62.5 & 47.7 & 41.5 & 44.4 & 51.8 & 48.5
   \\
   
   \quad + Label Distillation & \multirow{2}{*}{1402} & \multirow{2}{*}{37.5} & \multirow{2}{*}{22.5} & \multirow{2}{*}{40.4} & \multirow{2}{*}{62.5} & \multirow{2}{*}{44.4} & \multirow{2}{*}{60.4} &  \multirow{2}{*}{45.3} & \multirow{2}{*}{41.1} & \multirow{2}{*}{43.2} & \multirow{2}{*}{37.9} & \multirow{2}{*}{43.5} \\
   \quad \ \ \citep{wang-etal-2022-expanding} \\
   
    \hdashline\noalign{\vskip 0.5ex}
   \lexcgen-1K (\texttt{T}) & $\sim220$ & 17.8 & 27.9 & 29.4 & 34.8 & 31.0 & 24.9 & 29.8 & 28.6 & 29.2 & 29.8 & 28.3 \\
    \quad + Existing Task Data (\texttt{en}) & $\sim920$ & 31.8 & 37.8 & 37.3 & 65.0 & 50.0 & 59.7 & 46.8 & 35.9 & 37.9 & 48.1 & 45.0 
   \\
   
   \lexcgen-10K (\texttt{T}) & $\sim2.2$K & 39.3 & 40.3 & 50.0 & 64.2 & 55.9 & 66.5 & 55.0 & 41.4 & 46.5 & 54.9 & 51.4 \\
   \quad + Existing Task Data (\texttt{en}) & $\sim2.9$K & 36.9 & 42.4 & 50.6 & 67.2 & 55.9 & 64.8 & 54.6 & 39.8 & 46.4 & 53.9 & 51.2
   \\
   
   \textbf{\lexcgen-100K} (\texttt{T}) & $\sim22$K & \ul{48.4} & \ul{51.6} & \textbf{62.5} & \textbf{73.0} & \textbf{68.0} & \ul{70.3} & \ul{58.0} & \ul{41.7} & \textbf{53.7} & \textbf{62.7} & \ul{59.0}\\
   \textbf{\quad + Existing Task Data (\texttt{en})} & $\sim23$K & \textbf{48.6} & \textbf{53.6} & \textbf{62.5} & \ul{72.7} & \ul{65.2} & \textbf{72.8} & \textbf{60.3} & 41.2 & \ul{53.3} & \ul{61.7} & \textbf{59.2}
   \\

   \midrule
    \rowcolor{LightGray}\emph{Gold Translations} & 701 & 
    31.2 & 53.7 & 38.1 & 68.6 & 63.1 & 69.5 & 56.7 & 44.8 & 56.5 & 58.0 & 54.0 \\
   \bottomrule
   \end{tabular}
    \caption{Topic classification accuracy for 10 worst-performing languages in the SIB-200 dataset \cite{adelani2023sib200} with XLMR-base classifier \citep{conneau-etal-2020-unsupervised}. We follow the schema defined in Table~\ref{tab:sa-result}. 
    }
   \label{tab:tc-result-xlmr-base}
\end{table*}

\setul{1.5pt}{.4pt}
\begin{table*}[ht]
    \small
   \centering
   \begin{tabular}{lcccccccccccc}
   \toprule
   \textbf{Methods} & \textbf{\#data} & \textbf{bam} & \textbf{ewe} & \textbf{fij} & \textbf{grn} & \textbf{lin} & \textbf{lus} & \textbf{sag} & \textbf{tso} & \textbf{tum} & \textbf{twi} & \textbf{Avg }\\ 
   \midrule

   \emph{Zero-shot prompting}\\
   \midrule
   BLOOMZ-7.1.B & 0 & 41.7 & 34.3 & 35.3 & 41.7 & 42.2 & 38.7 & 36.8 & 41.7 & 40.2 & 41.7 & 39.4\\

    Aya-101-13B & 0 & 36.8 & 39.1 & 50.9 & 48.8 & 52.4 & 43.7 & 40.2 & \ul{54.1} & 50.0 & 37.7 & 45.4 \\
    Aya-101-13B (few-shot) & 5 & 42.2 & 46.1 & 60.4 & 55.1 & 59.7 & 48.2 & 49.4 & \textbf{56.2} & \textbf{57.5} & 43.8 & 51.9 \\
    \rowcolor{LightGray} GPT-4o & 0 & 58.1 & 56.2 & 63.9 & 75.8 & 69.4 & 65.3 & 57.8 & 57.2 & 59.8 & 64.8 & 67.7 \\
     \midrule
     \emph{Cross-lingual zero-shot} \\
     \midrule
   Existing Task Data (\texttt{en}) & 701 & 29.6 & 27.2 & 32.1 & 63.6 & 39.9 & 56.0 & 41.6 & 38.3 & 41.6 & 43.1 & 41.3 \\
    \midrule
   \emph{Word translation} \\
   \midrule
   Existing Task Data (\texttt{T}) & 701 & 42.4 & 43.1 & 48.5 & 70.6 & 52.9 & 66.4 & 43.4 & 43.5 & 47.7 & 52.9 & 51.1 \\

   \quad + Existing Task Data (\texttt{en}) & 1402 & 43.1 & 45.2 & 45.2 & 71.7 & 54.8 & 65.7 & 49.9 & 43.1 & 50.9 & 54.3 & 52.4 
   \\

   \quad + Label Distillation & \multirow{2}{*}{1402} & \multirow{2}{*}{37.9} & \multirow{2}{*}{27.8} & \multirow{2}{*}{42.9} & \multirow{2}{*}{64.6} & \multirow{2}{*}{43.5} & \multirow{2}{*}{58.9} &  \multirow{2}{*}{48.3} & \multirow{2}{*}{42.6} & \multirow{2}{*}{48.8} & \multirow{2}{*}{39.5} & \multirow{2}{*}{45.5} \\
   \quad \ \ \citep{wang-etal-2022-expanding} \\
   
    \hdashline\noalign{\vskip 0.5ex}
   \lexcgen-1K (\texttt{T}) & $\sim220$ & 23.5 & 32.4 & 33.9 & 47.1 & 35.3 & 44.7 & 34.1 & 27.2 & 33.0 & 26.2 & 33.7 \\
    \quad + Existing Task Data (\texttt{en}) & $\sim920$ & 37.5 & 45.7 & 41.8 & 70.2 & 52.8 & 60.7 & 48.2 & 43.3 & 44.6 & 51.0 & 49.6 
   \\
   
   \lexcgen-10K (\texttt{T}) & $\sim2.2$K & 43.2 & 46.6 & 53.3 & 68.1 & 59.1 & 68.1 & 50.6 & \ul{46.2} & \ul{55.5} & 53.2 & 54.4 \\
   \quad + Existing Task Data (\texttt{en}) & $\sim2.9$K & 37.5 & 44.3 & 51.7 & 69.2 & 57.7 & 68.1 & 49.6 & 42.4 & 51.3 & 58.2 & 53.0
   \\
   
   \textbf{\lexcgen-100K} (\texttt{T}) & $\sim22$K & \ul{50.5} & \textbf{54.6} & \ul{66.0} & \ul{74.1} & \textbf{67.5} & \textbf{70.7} & \ul{56.7} & 45.2 & \textbf{56.2} & \textbf{62.8} & \ul{60.4} \\
   \quad + Existing Task Data (\texttt{en}) & $\sim23$K & \textbf{52.4} & \ul{53.2} & \textbf{67.4} & \textbf{76.8} & \ul{67.0} & \ul{70.0} & \textbf{57.3} & 45.0 & 53.1 & \ul{62.5} & \textbf{60.5}
   \\

   \midrule
    \rowcolor{LightGray}\emph{Gold Translations} & 701 & 50.6 & 60.9 & 58.3 & 73.1 & 64.1 & 68.2 & 62.5 & 48.4 & 60.0 & 65.8 & 61.2 \\
   \bottomrule
   \end{tabular}
    \caption{Topic classification accuracy for 10 worst-performing languages in the SIB-200 dataset \cite{adelani2023sib200} with XLMR-large classifier \citep{conneau-etal-2020-unsupervised}. We follow the schema defined in Table~\ref{tab:sa-result}. 
    }
   \label{tab:tc-result-xlmr-large}
\end{table*}

\end{document}